\documentclass{article}

\usepackage[margin=1in]{geometry}   
\usepackage{times}                  
\usepackage{microtype}              
\usepackage{enumitem}               
\usepackage{caption}
\usepackage{subcaption}
\usepackage{booktabs}
\usepackage{graphicx}
\usepackage{amsmath,amssymb}
\usepackage{natbib}
\usepackage{hyperref}
\usepackage{xcolor}

\usepackage{float}

\usepackage{afterpage}
\usepackage{placeins}   

\usepackage{amsmath,amsfonts,bm}

\def\eqref#1{equation~\ref{#1}}

\def\1{\bm{1}}

\DeclareMathAlphabet{\mathsfit}{\encodingdefault}{\sfdefault}{m}{sl}
\SetMathAlphabet{\mathsfit}{bold}{\encodingdefault}{\sfdefault}{bx}{n}

\usepackage{hyperref}
\usepackage{url}

\usepackage{multirow}
\usepackage{algorithm}
\usepackage[noend]{algpseudocode}
\usepackage{xcolor}
\usepackage{float}
\usepackage{afterpage}

\usepackage[most]{tcolorbox}
\usepackage{enumitem}   
\usepackage{caption}    

\usepackage{graphicx,booktabs,makecell,array}
\newcommand{\imgh}{1.5cm} 
\newcommand{\thumb}[1]{\includegraphics[height=\imgh]{#1}}
\newcommand{\promptw}{0.25\linewidth} 
\newcommand{\promptcell}[1]{\parbox[c][\imgh][c]{\promptw}{\centering\tiny\vspace*{-5.3\baselineskip}#1}}
\newcommand{\editrow}[7]{%
  \thumb{#1} & \promptcell{#2} & \thumb{#3} & \thumb{#4} & \thumb{#5} & \thumb{#6} & \thumb{#7} \\[-14pt]
}
\newcommand{\genrow}[7]{%
  \promptcell{#1} & \thumb{#2} & \thumb{#3} & \thumb{#4} & \thumb{#5} & \thumb{#6} & \thumb{#7} \\[-14pt]
}

\newtcolorbox{promptbox}{
  enhanced, breakable,
  colback=white, colframe=black,
  arc=8pt, boxrule=0.9pt,
  left=10pt, right=10pt, top=8pt, bottom=8pt,
  fontupper=\ttfamily\small,      
}

\title{Image-POSER: Reflective RL for Multi-Expert Image Generation and Editing}

\author{
\begin{tabular}{cc}
$^{\dagger}$\textbf{Hossein Mohebbi}$^{1,2}$ &
$^{\dagger}$\textbf{Mohammed Abdulrahman}$^{1,2}$ \\
\small \texttt{mhmohebb@uwaterloo.ca} &
\small \texttt{m3abdulr@uwaterloo.ca} \\
\end{tabular}
\\[0.8em]
\begin{tabular}{ccc}
\textbf{Yanting Miao}$^{1,2}$ &
\textbf{Pascal Poupart}$^{1,2}$ &
\textbf{Suraj Kothawade}$^{3}$ \\
\small \texttt{y43miao@uwaterloo.ca} &
\small \texttt{ppoupart@uwaterloo.ca} &
\small \texttt{skothawade@google.com} \\
\end{tabular}
\\[0.8em]
$^1$University of Waterloo \quad
$^2$Vector Institute \quad
$^3$Google
\\[0.6em]
\small $^\dagger$Equal contribution
}

\begin{document}

\date{}  

\maketitle

\renewcommand{\abstractname}{\Large Abstract}
\patchcmd{\abstract}{\quotation}{\vspace{0.5em}\noindent}{}{}

\begin{abstract}
Recent advances in text-to-image generation have produced strong single-shot models, yet no individual system reliably executes the long, compositional prompts typical of creative workflows. 
We introduce \textbf{Image-POSER}, a \emph{reflective} reinforcement learning framework that (i) \textbf{orchestrates} a diverse registry of pretrained text-to-image and image-to-image experts, (ii) \textbf{handles long-form prompts end-to-end} through dynamic task decomposition, and (iii) \textbf{supervises alignment} at each step via structured feedback from a vision–language model critic. By casting image synthesis and editing as a Markov Decision Process, we learn non-trivial expert pipelines that adaptively combine strengths across models. Experiments show that Image-POSER outperforms baselines, including frontier models, across industry-standard and custom benchmarks in alignment, fidelity, and aesthetics, and is consistently preferred in human evaluations. These results highlight that reinforcement learning can endow AI systems with the capacity to autonomously decompose, reorder, and combine visual models, moving towards general-purpose visual assistants.
\end{abstract}

\vspace{-0.4cm}
\section{Introduction}
\label{intro}

Recent advances in image generation models such as GPT Image 1 and Gemini 2.5 Flash have made it possible to synthesize striking imagery from natural language prompts \citep{openai2025gptimage1, gemini2025flash}. Yet these systems often break down when confronted with \textbf{long, compositional instructions} that specify multiple objects, spatial relations, or sequential edits. For example, a marketing designer may request: ``\textit{Generate a product mockup with three bottles arranged diagonally on a wooden table, each with distinct labels, then restyle the background to match the brand’s colors.}” Today’s state-of-the-art models frequently miss such details: they miscount objects, ignore edits, or drift in style. Professionals are forced to manually stitch together a pipeline of specialized tools, iterating by trial and error until the output is acceptable. This workflow is inefficient, brittle, and inaccessible to non-experts who need precise, reliable results.

\begin{figure*}[t!]
\vspace{-0.3cm}
\centering
\includegraphics[width=0.7\textwidth]{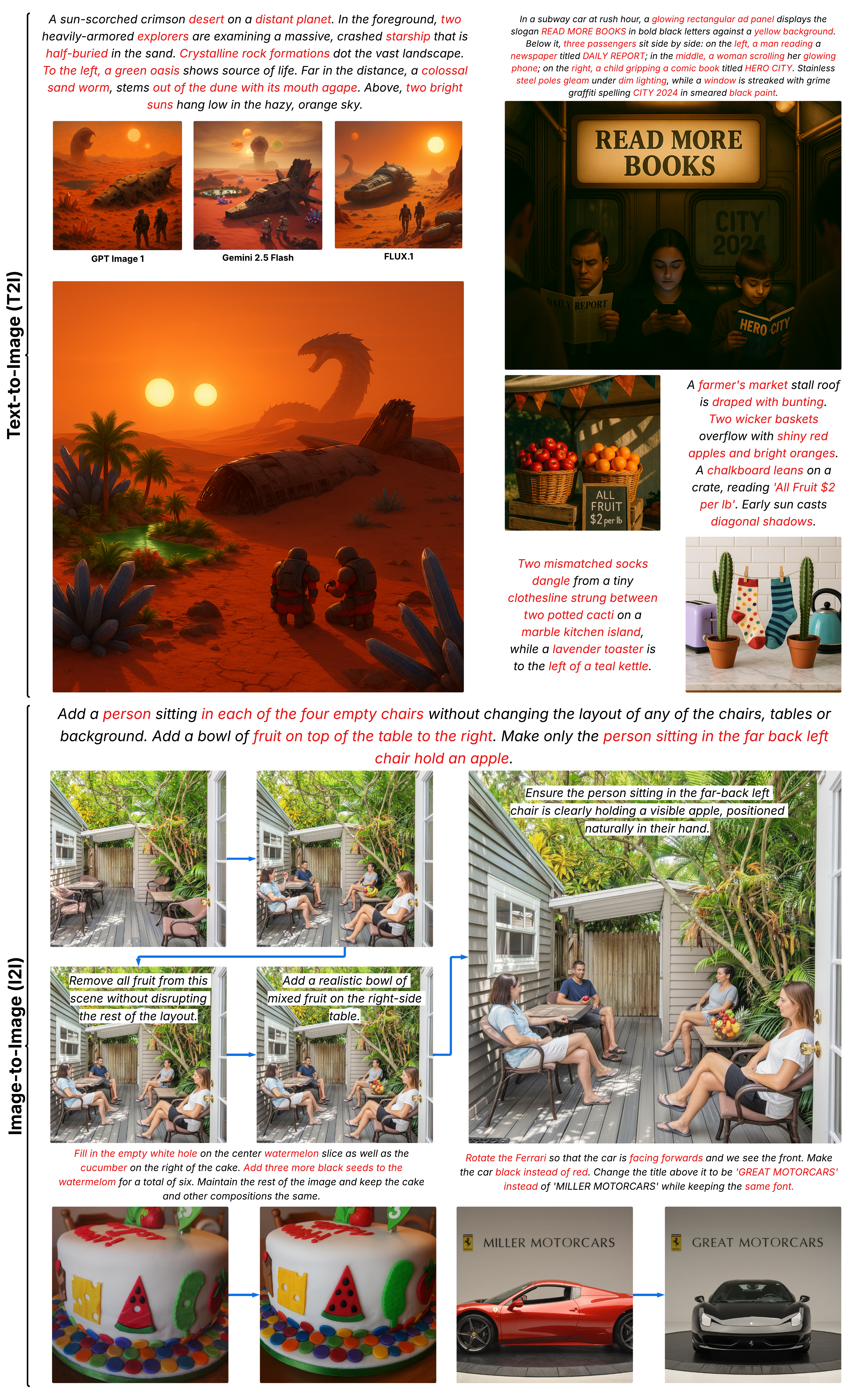}
\caption{\textbf{Select examples for complex long-form compositional prompts.} Top-Left: Text-to-Image (T2I) generations from multiple baselines versus Image-POSER, which successfully integrates all compositional constraints (object counts, spatial relations, style fidelity). Top-Right: additional T2I scenes with fine-grained requirements. Bottom: Image-to-Image (I2I) edits where Image-POSER completes multi-step instructions (adding/removing/counting objects, altering viewpoint, and preserving layout) that single-shot models struggle with. Images cited in Appendix~\ref{app:datasets}.}
\vspace*{-1.8em} 
\label{fig:hero}
\end{figure*}

A key limitation of current systems is the lack of \textbf{reflection and correction}. Human creators rarely succeed in one shot; they critique, refine, and retry. However, most models attempt a single forward pass \citep{podell2023sdxl, duan2025gotr1, chen2023pixartalpha}. When the result is misaligned, the user must intervene. This gap is especially costly in domains where \textbf{fidelity and consistency are non-negotiable}, such as product design, advertising, and content localization. Even small deviations, such as  incorrect object count, a missing logo, or a style mismatch, render outputs unusable.

We introduce \textbf{Image-POSER} (\underline{P}olicy-based \underline{O}rchestration for \underline{S}equential \underline{E}diting using \underline{R}eflection), a reinforcement learning (RL) framework that embeds reflection into the orchestration of visual experts. Starting from a prompt (and optionally an input image), Image-POSER proceeds in a loop: (i) a lightweight Deep Q-Network (DQN) agent selects an expert from a heterogeneous pool of text-to-image and image-to-image models; (ii) the chosen expert executes an instruction, producing an updated image; (iii) a vision–language model (VLM) critic evaluates the result, providing dense rewards and structured feedback that update the task set; and (iv) an auxiliary LLM module (Extract Command) isolates the next atomic instruction from the revised set, giving the agent a focused objective for the next step. By alternating between acting, critiquing, planning, and refining, Image-POSER can retry failed subtasks, adaptively re-plan, and compose non-trivial expert pipelines, thereby achieving reliability under complexity: the ability to faithfully execute intricate prompts that single-shot generation approaches fail to solve in one pass.

\textbf{Contributions} We present Image-POSER, a framework that formulates multi-expert image generation and editing as a sequential decision-making problem, enabling learned orchestration of diverse pretrained models. To support this, we design a reflective RL environment with two complementary modules: one that evaluates intermediate images and updates the set of remaining tasks, and another that extracts a single atomic task to guide the agent’s next action. We demonstrate that Image-POSER outperforms both single-model and agentic baselines on compositional benchmarks and user studies, achieving higher fidelity, alignment, and preservation on long-form prompts. By uniting reinforcement learning with image generation and reflective orchestration, Image-POSER moves toward generalist visual agents that can plan, execute, critique, and refine in ways that mirror how skilled artists combine brushes, layers, and filters to realize complex scenes.

\vspace{-0.3cm}
\section{Related Work}
\label{related}

Image-POSER is situated at the intersection of three active research directions: (i) orchestration of multiple pretrained visual experts for image generation and editing, (ii) reinforcement learning with model-based feedback, and (iii) reflection and task decomposition for compositional fidelity. While prior works have explored each of these threads separately, our approach unifies them into a single reflective RL framework.

\textbf{Image Generation and Editing} A survey of the image generation and editing models used as experts for Image-POSER can be found in Appendix~\ref{app:related_ref}.

\textbf{Orchestration of Multiple Visual Experts} Early work such as \emph{Visual ChatGPT} \citep{wu2023visual} demonstrated how a conversational LLM could call external vision foundation models to perform editing and generation, but relied heavily on user interaction and prompt chaining. \emph{GenArtist} \citep{wang2024genartist} extends this paradigm as a successor, replacing interactive dialogue with a multimodal LLM planner that decomposes prompts and executes a fixed schedule of expert calls (e.g., inpainting, style transfer). These approaches highlight the promise of expert orchestration but depend on heuristic planning or prompt engineering. In contrast, Image-POSER learns orchestration policies via reinforcement learning, enabling adaptive expert selection without manual scripting.

\textbf{Reinforcement Learning with Feedback for Text–Image Models} RL has also been applied to improve text-image alignment at the single-model level. Reinforcement Learning from AI Feedback (RLAIF) replaces costly human annotation with feedback from powerful LLM or VLM judges. \citet{xu2023imagereward} introduce \emph{ImageReward}, a reward model trained on human preferences that was later used in Reward Feedback Learning (ReFL) to fine-tune diffusion models for better prompt fidelity. These works validate LLM/VLM feedback as a scalable supervisory signal. Image-POSER builds on this principle but shifts the target of learning: instead of tuning one generator, we optimize an orchestration policy over a registry of experts.

\textbf{Reflection and Reasoning in Generation} A parallel line of work emphasizes reflection and decomposition as mechanisms to improve compositional fidelity. \emph{GoT-R1} \citep{duan2025got}, fine-tuned from \emph{Janus-Pro} \citep{chen2025janus}, rewards models for performing a ``Generation Chain-of-Thought'' that decomposes prompts into semantic and spatial components before synthesis. More recently, \citet{venkatesh2025crea} propose \emph{CREA}, a collaborative multi-agent framework that mirrors the human creative process through specialized roles such as a Creative Director, Prompt Architect, and Art Critic, coordinating to plan, critique, and refine creative outputs. CREA focuses on creativity-oriented reasoning but uses a fixed backbone, \emph{Flux.1-dev} for generation and \emph{ControlNet} for editing, so orchestration plays a limited role \citep{flux2024, zhang2023adding}. While these systems embed reflection within or across agents, Image-POSER introduces reflection at the orchestration level: a VLM critic evaluates intermediate results, updates residual tasks, and guides expert selection across a heterogeneous pool of experts. This enables retrying failed steps and adaptively refining plans across heterogeneous models, reframing the challenge from improving a single generator to learning how to compose multiple generators and editors.

Taken together, prior work demonstrates the value of orchestration, RL with feedback, and reflection, but each has remained siloed: orchestration has been heuristic, RL has been single-model, and reflection has been intra-model. Image-POSER integrates these threads into a unified reflective RL framework that learns to compose heterogeneous experts into effective pipelines, moving toward general-purpose visual assistants capable of planning, critiquing, and refining complex tasks.

\vspace{-0.2cm}
\section{Methodology}
\label{methodology}
\vspace{-0.2cm}

\begin{figure*}[t]
\vspace{-0.3cm}
\centering
\includegraphics[width=0.7\textwidth]{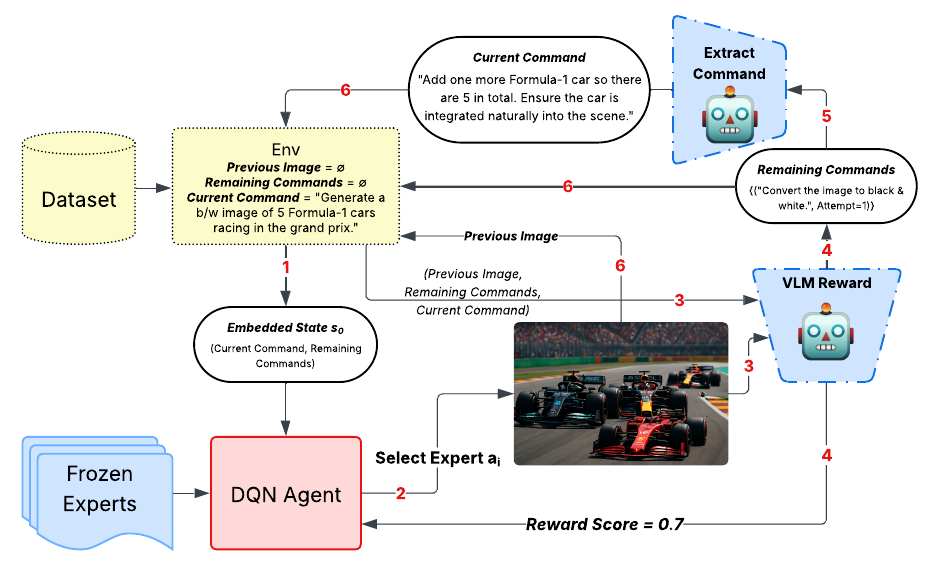} 
\caption{High-level \textit{example flow} of Image-POSER's pipeline for image generation and editing, numbered step by step. Illustrates the RL loop from the environment, to the DQN agent selecting a visual expert, to the VLM outputting a reward and reflecting for future tasks.}
\label{algorithm_flow}
\end{figure*}

We formulate multi-step image generation and editing as a sequential decision-making problem where an agent orchestrates a pool of pretrained visual experts. Rather than relying on a single model, the agent iteratively invokes text-to-image (T2I) or image-to-image (I2I) experts under reflective feedback. In the following formulation we focus on the input text-only setting, where the episode begins from a blank canvas. The extension to cases with an input image is straightforward: the loop simply begins from step $1$ with that image as the starting state. Each episode proceeds through a sequence of expert calls until the prompt is fulfilled or a step budget is reached. The overall process is detailed in Algorithm~\ref{alg:rl_pipeline} and illustrated in Figure~\ref{algorithm_flow}.

\subsection{Problem Formulation}

Given a textual prompt $\mathcal{P}$, the objective is to produce a final image $I$ that aligns with $\mathcal{P}$ in terms of content, spatial structure, and style. Unlike approaches that predefine a fixed sequence of subtasks, we treat the decomposition as \emph{dynamic and adaptive}: commands are incrementally created and revised as the system critiques intermediate outputs.

Let $\mathcal{E} = \{ e_1, \dots, e_N \}$ denote a registry of $N$ expert models, partitioned into text-to-image (T2I) and image-to-image (I2I) experts. The environment maintains two textual descriptors: the current command $c_{\text{curr}}$ and the set of remaining commands $C_{\text{rem}}$, where each element in $C_{\text{rem}}$ carries an attempt counter and is discarded by the reward model after three unsuccessful attempts. An episode produces a sequence ${I_0, I_1, \dots, I_T}$, where $I_0$ is either a blank canvas or an input image. At each step, the agent selects an action $a_t$ corresponding to an expert $e_{a_t} \in \mathcal{E}$, which is applied to $I_{t-1}$ under the instruction $c_{\text{curr}}$ to yield $I_t$. The environment then evaluates $I_t$, assigns a reward, and updates the command set.

An episode terminates when all aspects of the original prompt are satisfied or a maximum number of steps is reached. The agent’s objective is to maximize cumulative reward, thereby learning a policy that efficiently selects experts while maintaining alignment to the prompt.

\vspace{-0.15cm}
\subsection{State and Action Space}

\textbf{State} At time $t$, the RL state is a vector embedding
$
s_t = \text{embed}([c_{\text{curr}}, C_{\text{rem}}])
$
obtained from a pretrained text encoder. This embedding encodes both the immediate atomic command and a compressed summary of the outstanding commands.

\textbf{Action} The action space corresponds to the discrete set of experts in $\mathcal{E}$. To ensure validity, the action set is dynamically masked: only T2I experts are eligible when no image exists, and only I2I experts are eligible once an image has been generated. This prevents invalid calls, such as applying an image-conditioned model without an input image.

\vspace{-0.15cm}
\subsection{Reward Function and Reflection}

A central feature of Image-POSER is its reflective loop, composed of two complementary modules:

\textbf{Reward module (VLM critic)} After producing $I_t$, we query the VLM with $(I_{t-1}, I_t, c_{\text{curr}}, C_{\text{rem}}, \mathcal{P})$. The VLM evaluates alignment of $I_t$ with both the current command and the original prompt, returning a scalar score $r_t^{\text{raw}} \in [0,10]$ based on a rubric that asseses \textit{content accuracy}, \textit{spatial configuration}, \textit{visual quality} and \textit{style consistency}. Full rubric and system prompt are in Appendix \ref{app:system_prompts}.

The training reward is defined as
$
r_t = r_t^{\text{raw}}/10 - 0.05 t
$
which normalizes the score and penalizes longer pipelines.

To improve robustness, the critic also updates the remaining commands: if $c_{\text{curr}}$ is deemed incomplete, that is, the expert failed to fulfill it, it is returned to $C_{\text{rem}}$ with an incremented attempt counter. Any command that exceeds three attempts is permanently removed. This avoids excessive penalties for stochastic failures while preventing the agent from looping indefinitely.

\textbf{Extract command module (LLM)} Before the next step, a language model selects an atomic command from $C_{\text{rem}}$, giving priority to items with fewer attempts, and assigns it as $c_{\text{curr}}$. This ensures the agent always operates on a clear, well-formed instruction. See Appendix~\ref{app:system_prompts} for system prompt.

Together, these modules enable retries, refinements, and adaptive reordering of commands, allowing Image-POSER to construct non-trivial multi-step pipelines.

\vspace{-0.15cm}
\subsection{Learning Process}

The orchestration policy is learned via a Deep Q-Network (DQN). The Q-network is a lightweight multi-layer perceptron that maps the textual embedding $s_t$ to Q-values over the available experts. Training follows an $\epsilon$-greedy exploration strategy, with $\epsilon$ linearly annealed from 1.0 to 0.1 over 50\% of training horizon to balance exploration and exploitation.

Transitions $(s_t, a_t, r_t, s_{t+1}, \text{done})$ are stored in a replay buffer, and the Q-network is optimized with the standard DQN loss, minimizing the mean squared error between predicted Q-values and target values:
$\mathcal{L}(\theta) = \mathbb{E}\Bigg[ \Big( r_t + \gamma \max_{a'} Q_{\theta^-}(\mathbf{s}_{t+1}, a') - Q_\theta(\mathbf{s}_t, a_t) \Big)^2 \Bigg]$,  where $\theta^-$ are the parameters of a target network updated every 100 steps, and $\gamma = 0.99$ is the discount factor. Because the critic provides dense, normalized rewards with step penalties, the agent learns stable policies that favor concise and effective expert sequences.

\begin{algorithm}[t]
\caption{Image-POSER Pipeline for Multi-Step Image Generation \& Editing}
\label{alg:rl_pipeline}
\begin{small}
\begin{algorithmic}[1]
\State \textbf{Input:} Dataset of prompts $\mathcal{D}$, expert registry $\mathcal{E}$, max steps $T_{\max}=6$
\State Initialize DQN agent with Q-network $Q_\theta$, target network $Q_{\theta^-}$, replay buffer $\mathcal{B}$
\For{each training episode}
    \State Sample $(\mathcal{P}, I_0) \sim \mathcal{D}$
    \State Initialize $c_{\text{curr}} \gets \mathcal{P}$, $C_{\text{rem}} \gets \emptyset$, $s_0 \gets \texttt{embed}([c_{\text{curr}}, C_{\text{rem}}])$
    \For{$t = 1$ to $T_{\max}$}
        \State $a_t \gets \texttt{dqn.predict}(s_{t-1})$ \Comment{Filtered by input availability}
        \State $e_{a_t} \gets \mathcal{E}[a_t]$, $I_t \gets e_{a_t}(c_{\text{curr}}, I_{t-1})$
        \State $(r_t, C'_{\text{rem}}) \gets \texttt{vlm\_reward}(I_{t-1}, I_t, c_{\text{curr}}, C_{\text{rem}}, \mathcal{P})$
        \If{$c_{\text{curr}}$ is incomplete \textbf{and} attempt counter of $c_{\text{curr}} < 3$}
            \State Re-add $c_{\text{curr}}$ to $C'_{\text{rem}}$ with incremented attempt counter
        \EndIf
        \State $c_{\text{curr}} \gets \texttt{extract\_command}(C'_{\text{rem}})$, $C_{\text{rem}} \gets C'_{\text{rem}}$, $s_t \gets \texttt{embed}([c_{\text{curr}}, C_{\text{rem}}])$
        \State Store transition $(s_{t-1}, a_t, r_t, s_t, \text{done})$ in $\mathcal{B}$
        \State Sample batch from $\mathcal{B}$, update $Q_\theta$ via gradient descent
        \State Update $Q_{\theta^-}$ every 100 steps
        \If{$c_{\text{curr}} = \emptyset$ \textbf{and} $C_{\text{rem}} = \emptyset$} $done \gets True$, \textbf{break}
        \EndIf
    \EndFor
\EndFor
\end{algorithmic}
\end{small}
\end{algorithm}

\section{Experiment and Results}
\label{exp}
\vspace{-0.1cm}

\subsection{Experimental Setup}

\textbf{Expert Registry}
Our framework supports a heterogeneous set of image generation and editing models, unified under a common interface. For text-to-image (T2I) generation, we include \textit{Stable Diffusion XL} \citep{podell2023sdxl}, \textit{PixArt}-$\alpha$ \citep{chen2023pixartalpha}, \textit{Stable Diffusion 3.5 Large} \citep{stabilityai2024sd35blog}, \textit{DALL-E 3} \citep{openai2023dalle3}, \textit{GPT-Image-1} \citep{openai2025gptimage1}, \textit{FLUX.1-dev} \citep{flux2024}, and \textit{Gemini 2.5 Flash} \citep{gemini2025flash}. For image-to-image (I2I) editing, we support \textit{InstructPix2Pix} \citep{brooks2023instructpix2pix}, \textit{MagicBrush} \citep{zhang2023magicbrush}, \textit{FLUX Kontext} \citep{bfl2025fluxkontext}, \textit{GPT-Image-1 }\citep{openai2025gptimage1}, \textit{and Gemini 2.5 Flash} \citep{gemini2025flash}.

Each expert was evaluated individually to establish baseline performance. We also compare our system to other existing baselines designed for complex compositional prompts, including \emph{GoT-R1}~\citep{duan2025got} and \emph{GenArtist}~\citep{wang2024genartist}, which were not part of our registry. All baselines and experts were discussed in Section~\ref{related}.

\textbf{Datasets}
For training, we curated a dataset of 450 long-form prompts, combining human-authored and LLM-generated examples. These prompts emphasize multi-object compositions, spatial relations, and stylistic constraints. For evaluation, we considered three datasets: 1) \textbf{T2I-CompBench}~\citep{t2icompbench}: a widely used benchmark for compositional text-to-image generation, 2) \textbf{Custom T2I prompts}: 30 long-form generation prompts, designed to probe multi-step reasoning and compositional fidelity, and 3) \textbf{Custom I2I prompts}: 30 long-form editing prompts with paired input images, targeting object insertion, removal, and style preservation. See Appendix~\ref{app:datasets} for more details on the training and custom evaluation prompts.

\textbf{Implementation Details}
The orchestration agent is a lightweight DQN with a 3-layer MLP Q-network 
of dimensions $1536 \rightarrow 64 \rightarrow 64 \rightarrow 12$, 
mapping the text embedding state to Q-values over the expert set. For training our DQN, a single NVIDIA T4 GPU was used. For full details regarding our experimental setup, refer to Appendix~\ref{app:exp_details} and \ref{app:additional_training}.

\subsection{Quantitative Results}
\label{sec:quantitative}

We evaluate Image-POSER across both standardized benchmarks and custom-designed metrics. Three evaluation techniques are used:

\begin{enumerate}
    \item \textbf{T2I-CompBench}~\citep{t2icompbench} (Table~\ref{tab:t2i-compbench}), which measures compositional alignment using pretrained CLIP \citep{clip2021} and BLIP \citep{blip2022} scorers across attribute binding (color, shape, texture), object relationships (spatial, non-spatial), and complex composition.
    
    \item \textbf{CLIP/BLIP metrics} \citep{clip2021, blip2022} (Table~\ref{tab:blip_clip}), applied to the outputs of 30 long-form T2I prompts, where BLIP captures binding accuracy and CLIP captures non-spatial relational consistency.
    
    \item \textbf{GPT-o3 VLM judgments} \citep{openai2025o3} (Table~\ref{tab:gen-edit-results}), used to assess long-form generation and editing tasks. We used GPT-o3 with a modified system prompt that instructed it to output three reward dimensions: alignment, technical (fidelity), and aesthetics (for T2I), or alignment, preservation, and aesthetics (for I2I). This is the same VLM that powers Image-POSER’s reward loop, but with a different system prompt, provided in Appendix~\ref{app:system_prompts}.
\end{enumerate}

As shown in Table~\ref{tab:t2i-compbench}, Image-POSER achieves the strongest scores across all attribute binding categories and outperforms baselines in spatial reasoning. On long-form T2I prompts (Tables~\ref{tab:gen-edit-results} and \ref{tab:blip_clip}), Image-POSER surpasses baselines across both CLIP/BLIP metrics and GPT-o3 VLM judgments, with especially large gains in alignment and technical fidelity. For I2I editing prompts (Table~\ref{tab:gen-edit-results}), Image-POSER achieves the best overall performance, combining high alignment and aesthetics with competitive preservation scores.

In Table~\ref{tab:gen-edit-results}, Wilcoxon signed-rank tests \citep{wilcoxon1945individual} confirm that Image-POSER’s improvements over the baselines are statistically significant ($\downarrow$ when $p < 0.05$ and $\Downarrow$ when $p < 0.01$). The only case where a baseline outperformed Image-POSER numerically was in I2I Preservation, but the difference was not statistically significant. These results highlight the robustness of reflective orchestration: by adaptively sequencing expert calls, Image-POSER consistently delivers higher-fidelity outputs across both generation and editing.

\begin{table*}[b]
\caption{\textbf{T2I-CompBench category scores (higher is better)}. Image-POSER achieves the strongest results across all attribute binding dimensions (color, shape, texture) and outperforms baselines on spatial reasoning. Note * indicates results were sourced from original paper.}

\label{tab:t2i-compbench}
\centering
\scriptsize
\setlength{\tabcolsep}{10pt}
\renewcommand{\arraystretch}{1.05}

\begin{tabular}{lcccccc}
\toprule
& \multicolumn{3}{c}{\textbf{Attribute Binding}} & \multicolumn{2}{c}{\textbf{Object Relationship}} & \multirow{2}{*}{\textbf{Complex}} \\
\cmidrule(lr){2-4} \cmidrule(lr){5-6}
\textbf{Method} & \textbf{Color} & \textbf{Shape} & \textbf{Texture} & \textbf{Spatial} & \textbf{Non-Spatial} & \\
\midrule
*SD XL \citep{podell2023sdxl}   & 0.5879 & 0.4687 & 0.5299 & 0.2133 & 0.3119 & 0.3237 \\
*DALL-E 3 \citep{openai2023dalle3}              & 0.7785 & 0.6205 & 0.7036 & 0.2865 & 0.3003 & 0.3773 \\
*FLUX.1 \citep{flux2024}                & 0.7407 & 0.5718 & 0.6922 & 0.2863 & 0.3127 & 0.3703 \\
*GoT-R1-7B \citep{duan2025gotr1}             & 0.8139 & 0.5549 & 0.7339 & 0.3306 & 0.3169 & \textbf{0.3944} \\
*PixArt-$\alpha$ \citep{chen2023pixartalpha}       & 0.6690 & 0.4927 & 0.6477 & 0.2064 & \textbf{0.3197} & 0.3433 \\
GenArtist \citep{wang2024genartist}             & 0.4775 & 0.4491 & 0.5113 & 0.1587 & 0.2953 & 0.3073 \\
GPT Image 1 \citep{openai2025gptimage1}           & 0.8253 & 0.6145 & 0.7406 & 0.4218 & 0.3158 & 0.3732 \\
Gemini 2.5 Flash \citep{gemini2025flash}                & 0.8188 & 0.5572 & 0.7089 & 0.2853 & 0.3069 & 0.3537 \\
\textit{Image-POSER (ours)}       & \textbf{0.8572} & \textbf{0.6218} & \textbf{0.7595} & \textbf{0.4440} & 0.3195 & 0.3832 \\
\bottomrule
\end{tabular}
\end{table*}

\begin{table*}[t]
\caption{\textbf{Quantitative comparison of Image-POSER and baselines on complex long-form T2I/I2I prompts.} Evaluated using a VLM as a judge across 3 key dimensions (higher is better). Arrows indicate statistical significance against Image-POSER under the Wilcoxon signed-rank test. Note that $\downarrow$ ($p < 0.05$) and $\Downarrow$ ($p < 0.01$) indicate Image-POSER performs significantly better than the baseline, while $\uparrow$ and $\Uparrow$ denote the same significance levels when the baseline performs significantly better than Image-POSER.}
\label{tab:gen-edit-results}
\centering
\scriptsize
\setlength{\tabcolsep}{10pt}
\renewcommand{\arraystretch}{1.05}

\begin{tabular}{lcccc}
\toprule

\textbf{Generation Methods} & \textbf{Alignment} & \textbf{Technical} & \textbf{Aesthetic} & 
\textbf{Average}
\\
\midrule

SD 3.5 Large \citep{stabilityai2024sd35blog} & 78.61 $\pm$ 3.21 $\Downarrow$ & 93.23 $\pm$ 0.68 $\Downarrow$ & 88.67 $\pm$ 1.33 $\downarrow$ & 86.84 $\pm$ 1.33 $\Downarrow$ \\
SD XL \citep{podell2023sdxl} & 54.29 $\pm$ 3.48 $\Downarrow$ & 91.50 $\pm$ 1.60 $\Downarrow$ & 80.00 $\pm$ 2.25 $\Downarrow$ & 75.26 $\pm$ 2.21 $\Downarrow$ \\
DALL-E 3 \citep{openai2023dalle3} & 77.43 $\pm$ 3.12 $\Downarrow$ & 94.13 $\pm$ 0.85 $\Downarrow$ & 90.00 $\pm$ 0.00 $\downarrow$ & 87.19 $\pm$ 1.31 $\Downarrow$ \\
FLUX.1 \citep{flux2024} & 81.56 $\pm$ 2.94 $\Downarrow$ & 95.87 $\pm$ 0.87 $\downarrow$ & 89.60 $\pm$ 0.76 $\downarrow$ & 89.01 $\pm$ 1.21 $\Downarrow$ \\
GoT-R1-7B \citep{duan2025gotr1} & 75.70 $\pm$ 3.14 $\Downarrow$ & 91.13 $\pm$ 0.87 $\Downarrow$ & 84.67 $\pm$ 2.08 $\Downarrow$ & 83.83 $\pm$ 1.44 $\Downarrow$ \\
PixArt-$\alpha$ \citep{chen2023pixartalpha} & 51.19 $\pm$ 3.51 $\Downarrow$ & 92.60 $\pm$ 1.05 $\Downarrow$ & 87.20 $\pm$ 1.26 $\Downarrow$ & 77.00 $\pm$ 2.33 $\Downarrow$ \\
GenArtist \citep{wang2024genartist} & 46.37 $\pm$ 3.53 $\Downarrow$ & 82.53 $\pm$ 1.61 $\Downarrow$ & 61.33 $\pm$ 3.24 $\Downarrow$ & 63.41 $\pm$ 2.29 $\Downarrow$ \\
GPT Image 1 \citep{openai2025gptimage1} & 93.80 $\pm$ 1.81 $\downarrow$ & 96.93 $\pm$ 0.65  & 90.40 $\pm$ 0.59  & 93.71 $\pm$ 0.72 $\Downarrow$ \\
Gemini 2.5 Flash \citep{gemini2025flash} & 92.12 $\pm$ 1.73 $\Downarrow$ & 95.07 $\pm$ 0.92 $\Downarrow$ & 89.33 $\pm$ 0.95 $\downarrow$ & 92.17 $\pm$ 0.76 $\Downarrow$ \\
\textit{Image-POSER (ours)} & \textbf{96.65 $\pm$ 1.01} & \textbf{97.57 $\pm$ 0.64} & \textbf{91.33 $\pm$ 0.63} & \textbf{95.18 $\pm$ 0.53} \\

\midrule

\textbf{Editing Methods} & \textbf{Alignment} & \textbf{Preservation} & \textbf{Aesthetic} & 
\textbf{Average}
\\

\midrule

MagicBrush \citep{zhang2023magicbrush} & 45.16 $\pm$ 5.14 $\Downarrow$ & \textbf{80.93 $\pm$ 3.12} & 70.83 $\pm$ 3.07 $\Downarrow$ & 65.64 $\pm$ 2.74 $\Downarrow$ \\
InstructPix2Pix \citep{brooks2023instructpix2pix} & 33.55 $\pm$ 5.02 $\Downarrow$ & 74.50 $\pm$ 5.02 $\downarrow$ & 62.27 $\pm$ 3.46 $\Downarrow$ & 56.77 $\pm$ 3.18 $\Downarrow$ \\
FLUX Kontext \citep{bfl2025fluxkontext} & 78.98 $\pm$ 2.82 $\Downarrow$ & 80.40 $\pm$ 5.08 & 82.67 $\pm$ 2.03 $\Downarrow$ & 80.68 $\pm$ 2.03 $\Downarrow$ \\
GPT Image 1 \citep{openai2025gptimage1} & 91.34 $\pm$ 1.90 $\downarrow$ & 70.27 $\pm$ 6.96 $\downarrow$ & 89.13 $\pm$ 0.92 $\downarrow$ & 83.58 $\pm$ 2.60 $\Downarrow$ \\
Gemini 2.5 Flash \citep{gemini2025flash} & 87.79 $\pm$ 2.90 $\downarrow$ & 74.43 $\pm$ 6.96  & 86.43 $\pm$ 1.64 $\Downarrow$ & 82.89 $\pm$ 2.62 $\downarrow$ \\
\textit{Image-POSER (ours)} & \textbf{94.26 $\pm$ 1.46} & 80.13 $\pm$ 5.87 & \textbf{91.07 $\pm$ 1.00} & \textbf{88.49 $\pm$ 2.12} \\

\bottomrule
\end{tabular}
\end{table*}

\begin{figure*}[h!]
\centering
\setlength{\tabcolsep}{2pt}
\renewcommand{\arraystretch}{0.95}
\begin{tabular}{@{}c c c c c c c@{}}
\tiny\textbf{Prompt} &
\tiny\makecell{\textbf{PixArt-$\alpha$}} &
\tiny\makecell{\textbf{SD 3.5 Large}} &
\tiny\makecell{\textbf{FLUX.1-dev}} &
\tiny\makecell{\textbf{Gemini 2.5 Flash}} &
\tiny\makecell{\textbf{GPT Image 1}} &
\tiny\textbf{Image-POSER} \\
\midrule
\genrow
{A basketball court in the middle of New York City. \textcolor{red}{Six players} are playing 3v3. There are also \textcolor{red}{two spectators watching. One player is mid-dunk} in this sequence.}
{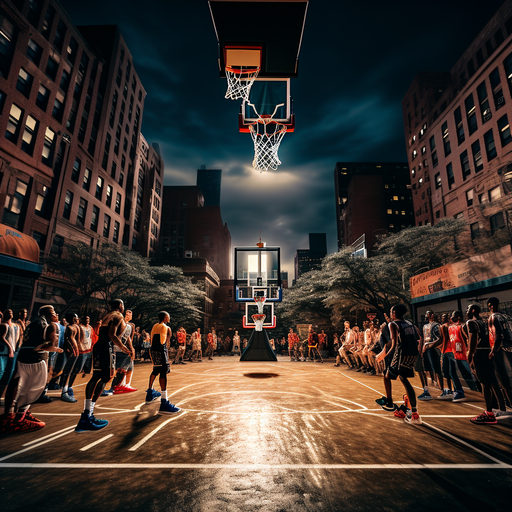}
{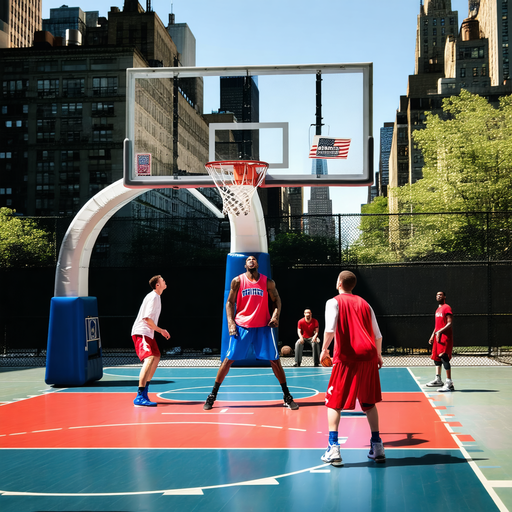}
{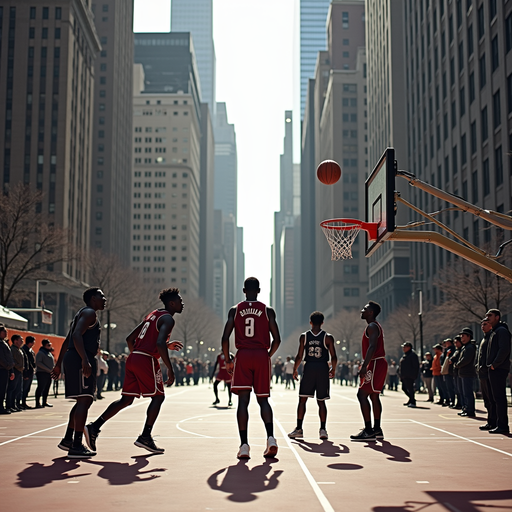}
{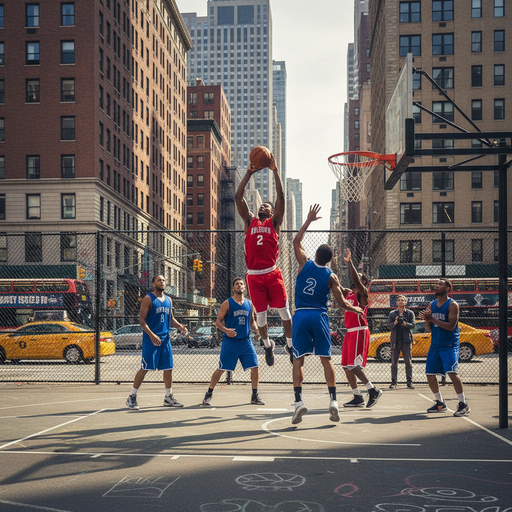}
{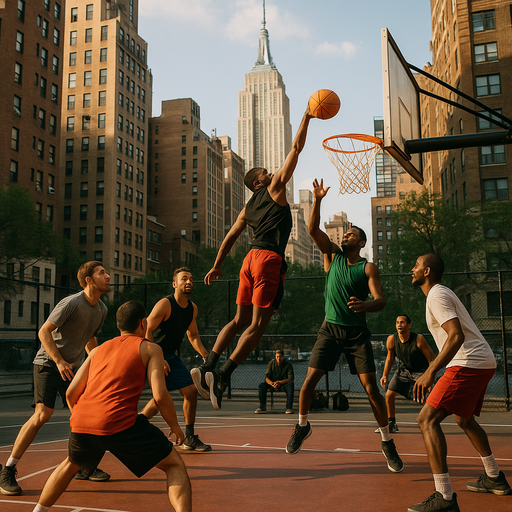}
{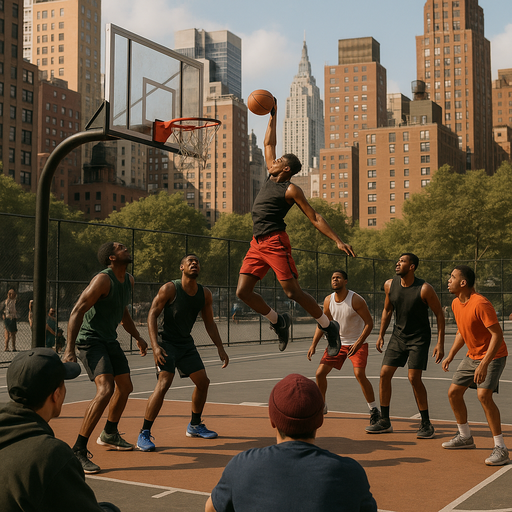}
\genrow
{A soccer field under a bright noon sun. The \textcolor{red}{scoreboard in the back reads 'HOME 3 : AWAY 1' in amber LEDs.} A midfielder in red jersey \#7 is mid-kick, right leg fully extended; a defender in blue jersey slides in from the lower left.}
{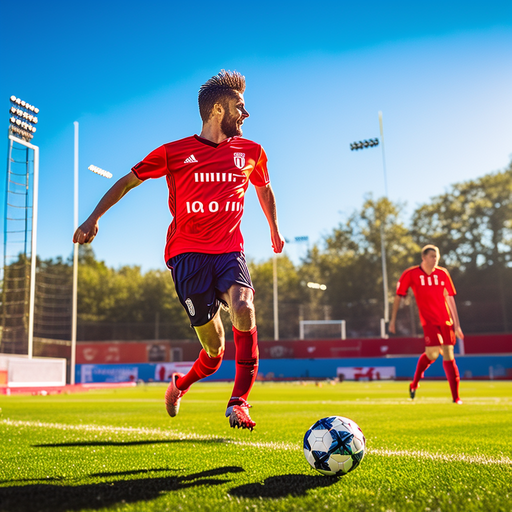}
{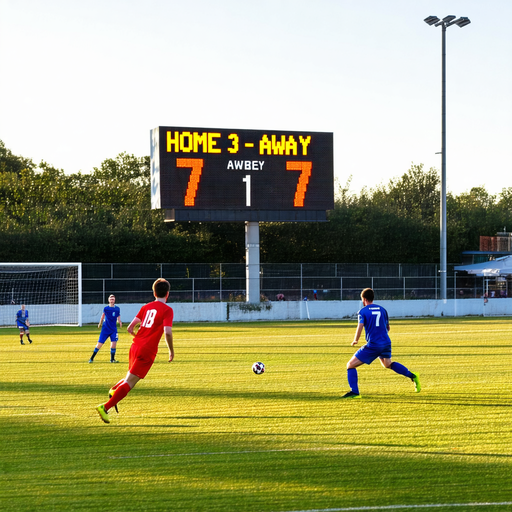}
{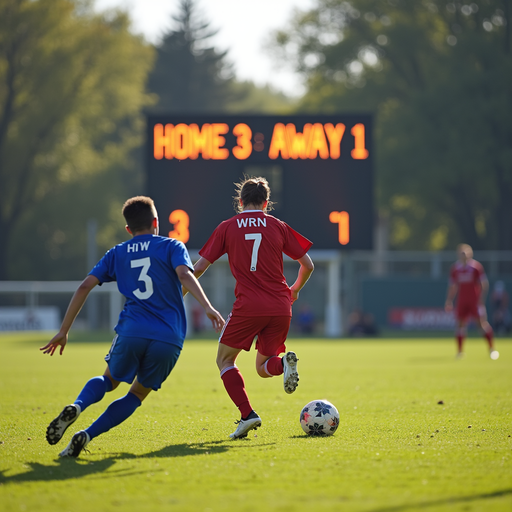}
{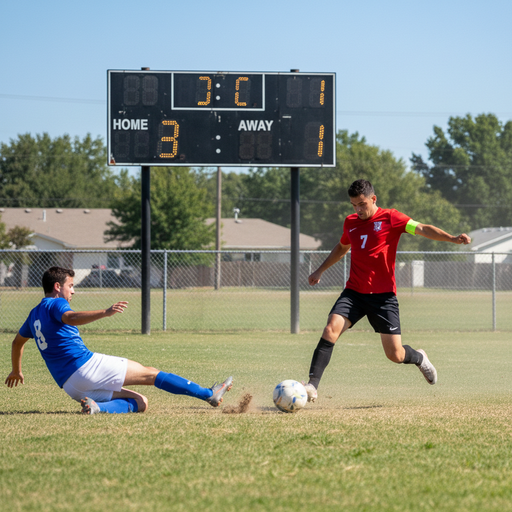}
{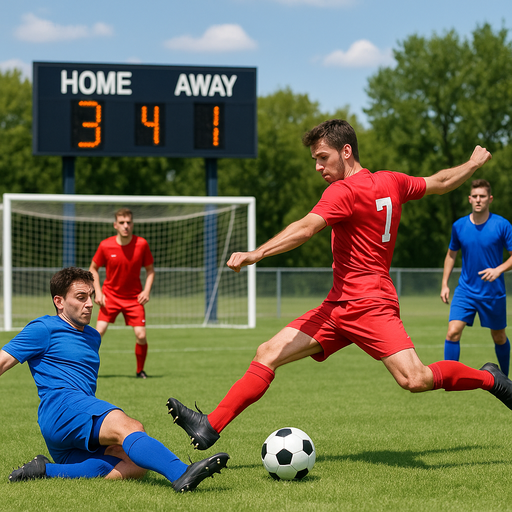}
{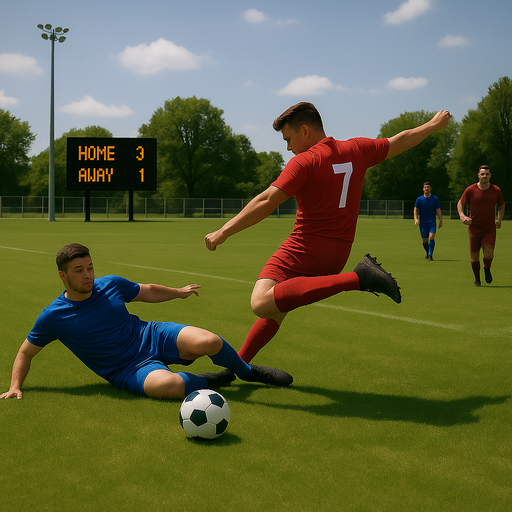}
\genrow
{\textcolor{red}{Six paper boats} float in a baking tray full of marbles, while a red pepper mill \textcolor{red}{to the right of} a blue mug watches the regatta.}
{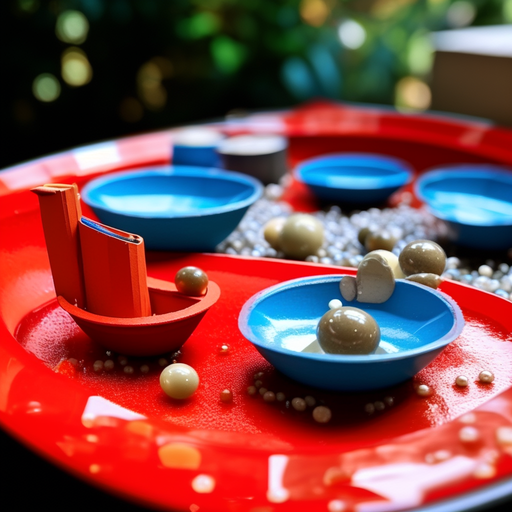}
{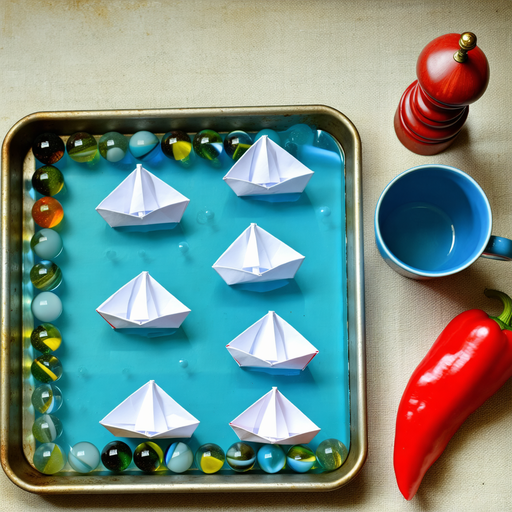}
{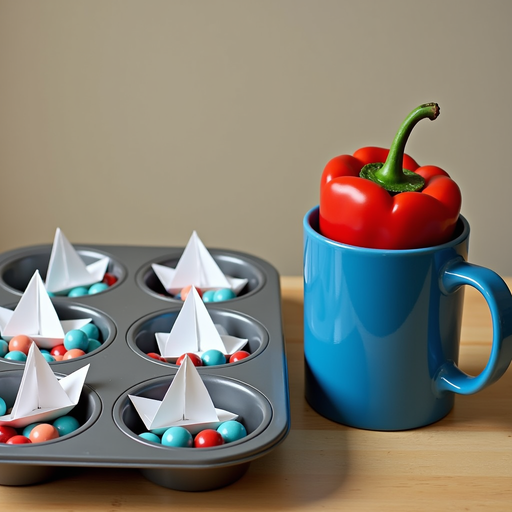}
{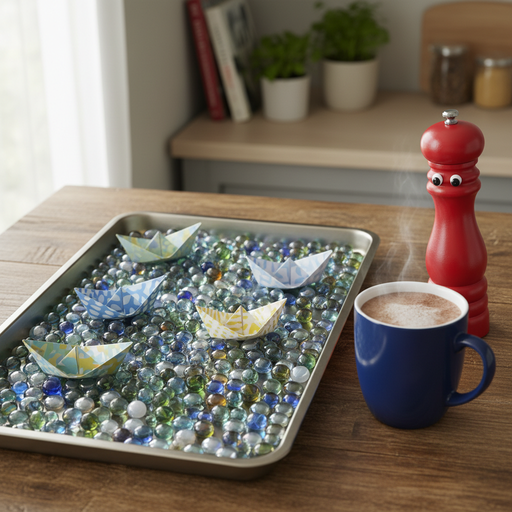}
{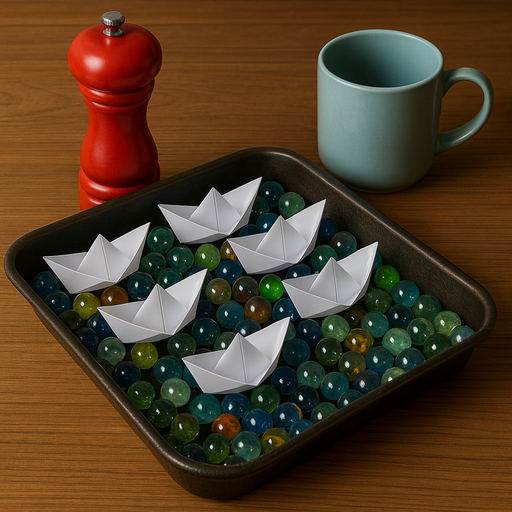}
{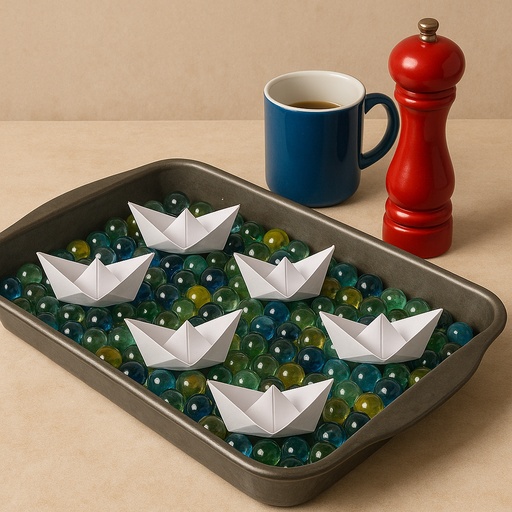}
\end{tabular}

\vspace{-3mm}
\end{figure*}

\begin{figure*}[h!]
\centering
\setlength{\tabcolsep}{2pt}
\renewcommand{\arraystretch}{0.95}
\begin{tabular}{@{}c c c c c c c@{}}
\tiny\makecell{\textbf{Input Image}} &
\tiny\textbf{Instruction} &
\tiny\makecell{\textbf{MagicBrush}} &
\tiny\makecell{\textbf{FLUX Kontext}} &
\tiny\makecell{\textbf{Gemini 2.5 Flash}} &
\tiny\makecell{\textbf{GPT Image 1}} &
\tiny\textbf{Image-POSER} \\
\midrule
\editrow{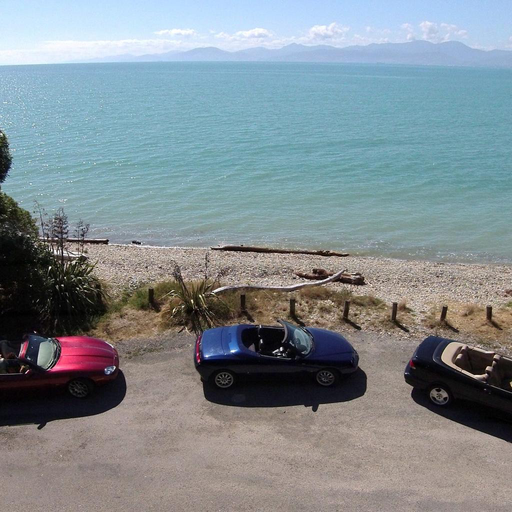}
{\textcolor{red}{Change the colors of the cars} in the middle and on the right to match the car on the left. \textcolor{red}{Add six sailboats} out in the distant ocean background.}
{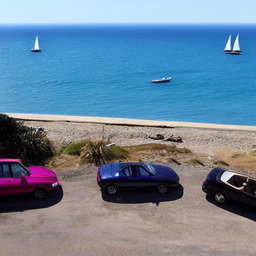}
{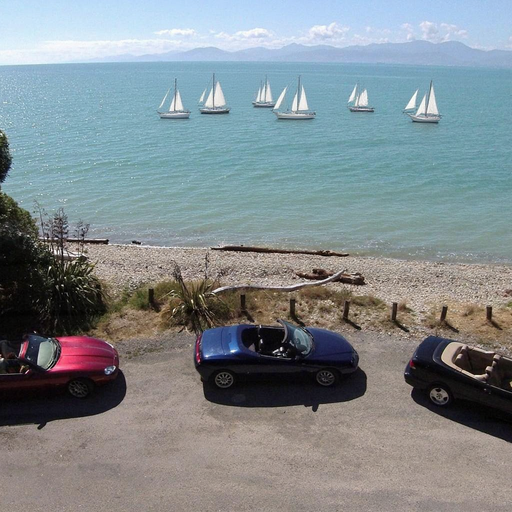}
{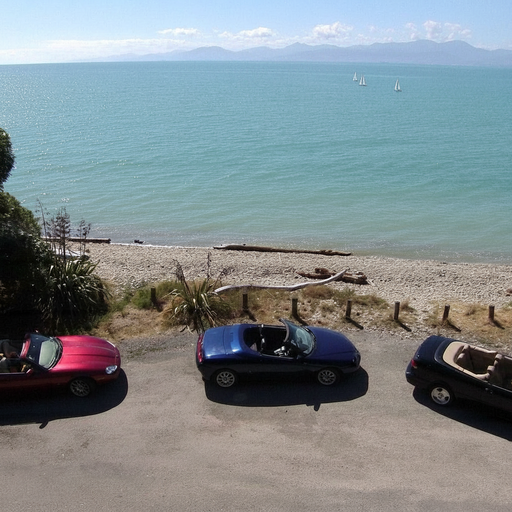}
{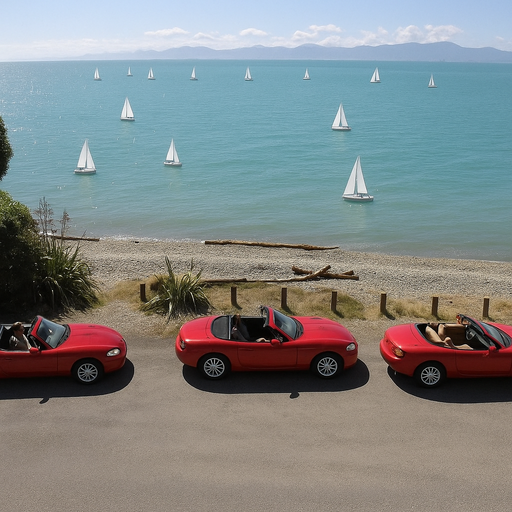}
{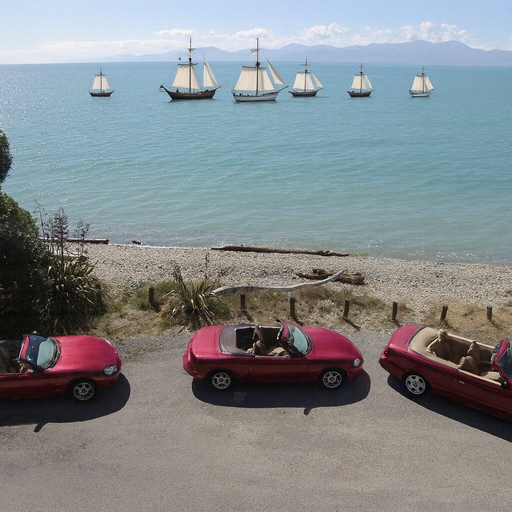}
\editrow{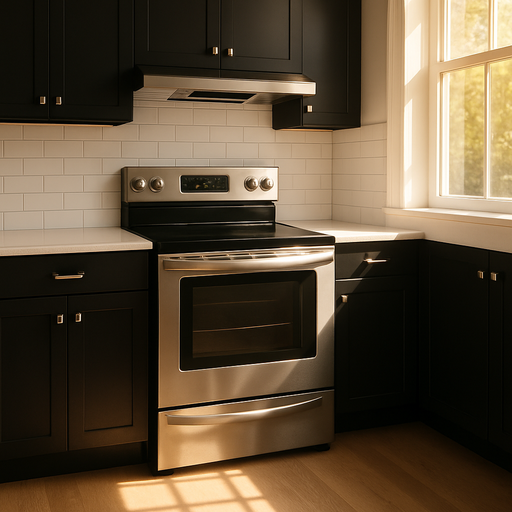}
{Change the color of the kitchen cabinets to be vintage green. Add a stuffed turkey cooking inside the oven visible from the window. \textcolor{red}{Add a stack of exactly three pots} on the kitchen countertop and \textcolor{red}{one single frying pan} on the stove.}
{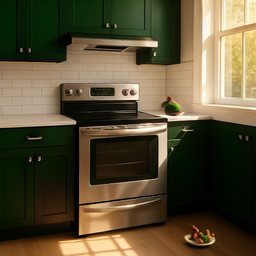}
{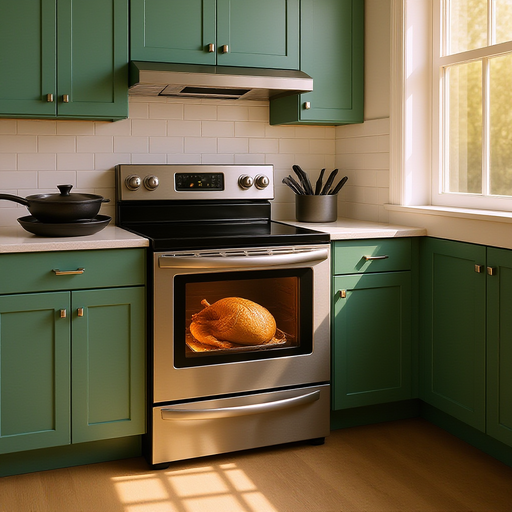}
{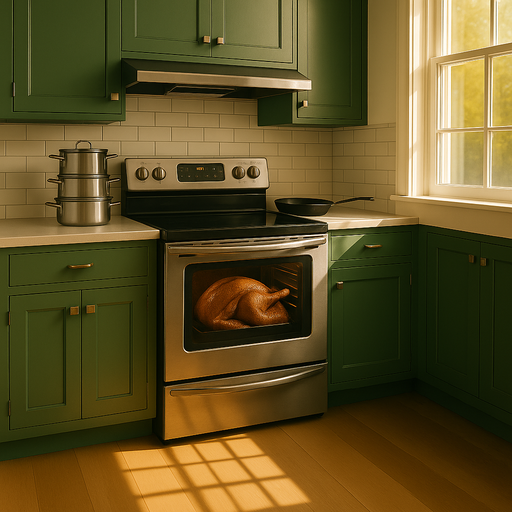}
{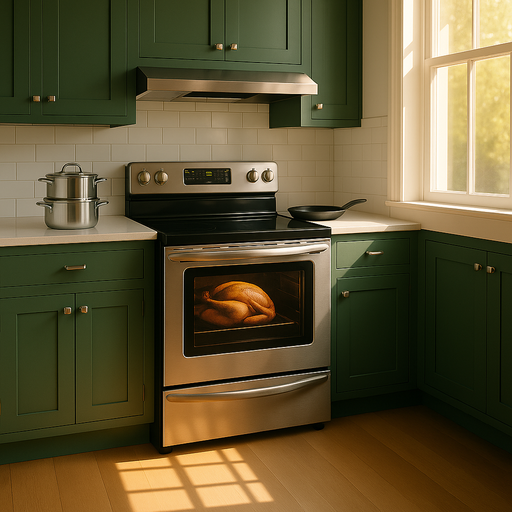}
{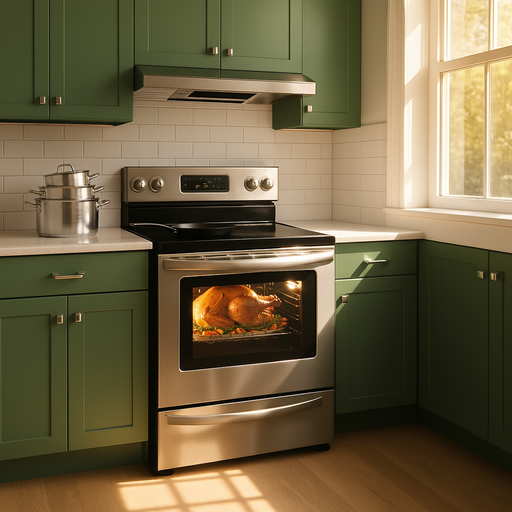}
\editrow{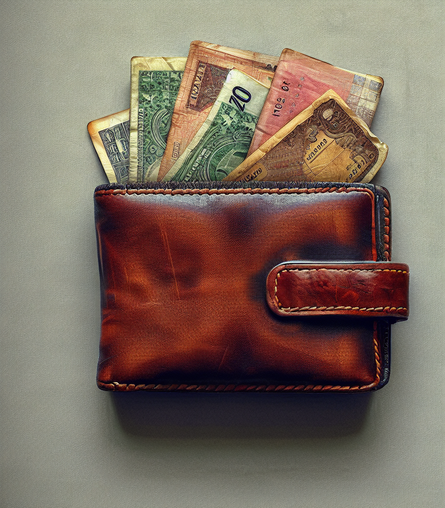}
{Apply an \textcolor{red}{animated style} to this image. Remove the existing bills and replace them with three new bills poking out of the wallet. The bills should read, \textcolor{red}{`\$100', `\$20', and `\$50' in that sequence}. All the bills should be green.}
{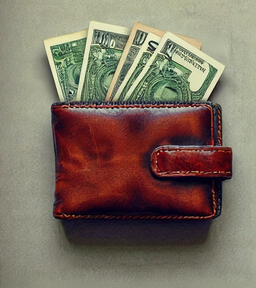}
{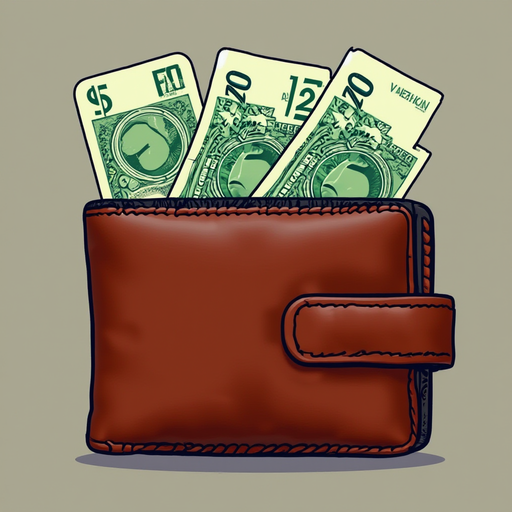}
{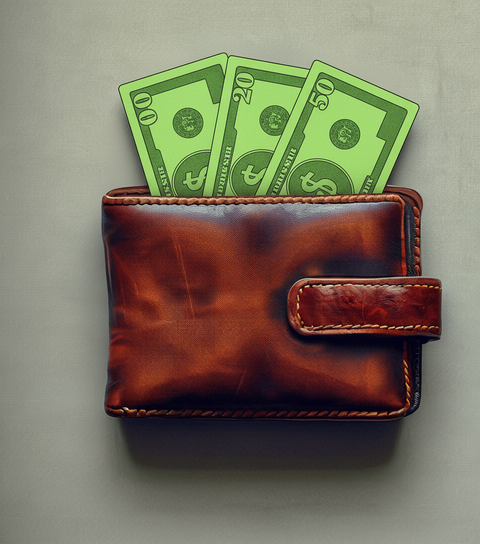}
{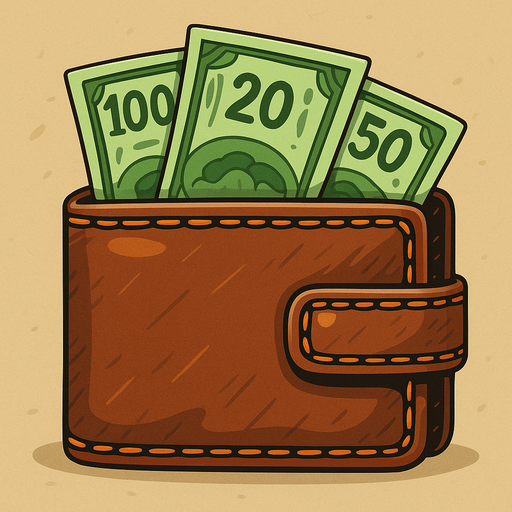}
{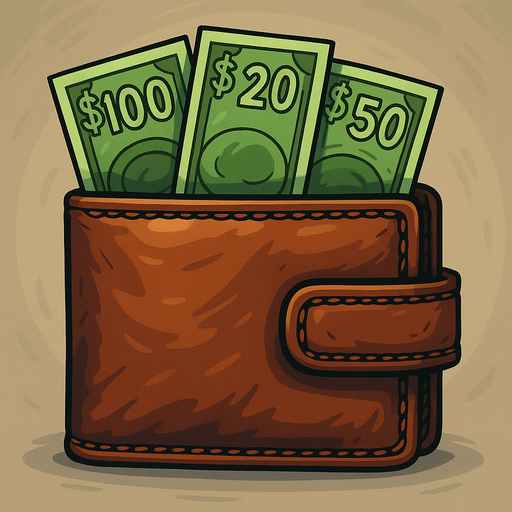}
\end{tabular}
\caption{\textbf{Qualitative comparison of long-form prompts for generation (top) and editing (bottom).} Baselines often fail on compositional constraints such as object counts, spatial relations, and object addition/removal. Image-POSER produces accurate, context-aware outputs that align with the instructions.}
\label{fig:gen-edit-comparison}
\vspace{-3mm}
\end{figure*}

\subsection{Qualitative Results}
\label{sec:qualitative}
\vspace{-0.2cm}

Figures~\ref{fig:hero} and \ref{fig:gen-edit-comparison} illustrate qualitative comparisons. They highlight Image-POSER’s ability to follow long instructions requiring multiple refinements, such as inserting objects while preserving style or applying sequential edits. In generation tasks, Image-POSER captures both local details (e.g., correct number of players in a sports scene) and global compositional structure (e.g., spatial layout, perspective). In editing tasks, Image-POSER produces accurate, context-aware modifications (e.g., changing object counts, applying style constraints) without degrading unrelated regions.

These examples emphasize the need for multi-step refinement: while strong baselines can produce aesthetic single-shot generations, they often miss compositional constraints. In contrast, Image-POSER incrementally corrects failures, yielding outputs that are both faithful and visually coherent.

\vspace{-0.15cm}
\subsection{User Study}
\label{sec:userstudy}
\vspace{-0.25cm}

To complement the automated metrics, we conducted a human preference study. For each setting (generation and editing), 30 prompts were sampled, and participants were shown side-by-side outputs: one from Image-POSER and one from a baseline model (while randomizing the order and hiding which model produced which image). In total, 14 volunteers completed the surveys, and care was taken to ensure that no participant ever saw the same image twice.

Results are presented in Table~\ref{tab:userstudy}. In T2I generation, Image-POSER outperformed every baseline, with especially large margins over strong systems such as GPT-Image~1 and Gemini~2.5 Flash, where it more reliably satisfied compositional requirements. In I2I editing, Image-POSER again won against all baselines, including notable gains over MagicBrush and FLUX Kontext, which surpassed Image-POSER on the Preservation dimension in Table~\ref{tab:gen-edit-results}. Across both tasks, annotators consistently favored Image-POSER for alignment and technical fidelity.

Interestingly, we observed a reversal in relative rankings between Gemini and GPT-Image~1. While GPT-Image~1 generally outperformed Gemini in our automated metrics (Tables~\ref{tab:t2i-compbench}, \ref{tab:gen-edit-results}, \ref{tab:blip_clip}), Gemini received higher preference in the user study. We attribute this to Gemini’s strong rendering fidelity: its images often appear more natural and less overtly “AI-generated,” which can make them more appealing in human evaluations even when compositional alignment is imperfect. Importantly, Image-POSER surpassed both models in both studies, indicating that reflective orchestration improves not only automated alignment scores but also human-perceived quality.

\begin{table*}[b]
\centering
\caption{\textbf{Win Rates from User Study.} Table shows the average rate at which annotators preferred Image-POSER’s outputs over a given baseline (higher is better for Image-POSER).}
\label{tab:userstudy}
\scriptsize
\setlength{\tabcolsep}{1pt}
\renewcommand{\arraystretch}{1.05}

\begin{tabular}{ccccccccc}
\toprule
& & & & \textbf{\textit{Text-to-Image}} \\
\midrule

\textbf{SD 3.5 Large} & \textbf{SD XL} & \textbf{DALL-E 3} & \textbf{FLUX.1} & \textbf{GoT-R1-7B} & \textbf{PixArt-$\alpha$} & \textbf{GenArtist} & \textbf{GPT Image 1} & \textbf{Gemini 2.5 Flash} 

\\

0.80 $\pm$ 0.07 & 
0.97 $\pm$ 0.03 & 
0.80 $\pm$ 0.07 & 
0.80 $\pm$ 0.07 & 
0.97 $\pm$ 0.03 & 
0.93 $\pm$ 0.05 & 
0.93 $\pm$ 0.05 & 
0.67 $\pm$ 0.09 & 
0.57 $\pm$ 0.09 
\\
\midrule

& & & & \textbf{\textit{Image-to-Image}} \\
\midrule
& & 
\textbf{MagicBrush} & \textbf{InstructPix2Pix} & \textbf{FLUX Kontext} & \textbf{GPT Image 1} & \textbf{Gemini 2.5 Flash} 

\\
& & 
0.97 $\pm$ 0.03 & 
0.97 $\pm$ 0.03 & 
0.80 $\pm$ 0.07 & 
0.67 $\pm$ 0.09 & 
0.60 $\pm$ 0.09 
\\

\bottomrule
\end{tabular}
\end{table*}

\section{Discussion}
\label{sec:discussion}

\subsection{Reflection on Image-POSER’s Contributions}
\vspace{-0.05cm}

Image-POSER demonstrates that \emph{reflective orchestration} can consistently outperform monolithic generators on complex compositional tasks. By combining retries, dynamic decomposition, and adaptive sequencing of experts, the framework remains robust across diverse prompt types.

A closer analysis clarifies why reflective orchestration is necessary. Figure~\ref{fig:avg-scores} reports the average reward scores assigned to individual experts, showing that while some models are consistently stronger overall (e.g., GPT Image 1, Gemini), others lag behind. Yet Figure~\ref{fig:task-distribution} reveals that no single model dominates across all task types: one expert may excel in object addition while another performs better in object resizing or background replacement. Task categories were automatically annotated by the \textit{extract command} module, which, alongside extracting $c_{\text{curr}}$ and $C_{\text{rem}}$, assigned each command to a taxonomy of editing operations (e.g., object addition, removal, resizing). This reveals a central challenge: top-performing models still show significant variability across task types.

The same pattern is evident in the quantitative editing results (Table~\ref{tab:gen-edit-results}). Image-POSER achieves overall state-of-the-art performance not by uniformly surpassing every expert, but by combining their complementary strengths. For instance, MagicBrush and FLUX Kontext achieve higher Preservation scores than frontier models such as GPT-Image-1 or Gemini. However, Image-POSER’s preservation performance tracks much closer to MagicBrush and FLUX Kontext than to the frontier models, indicating that it inherits these strengths through orchestration while still maintaining top-tier alignment overall. In practice, reflective orchestration operates as a mixture-of-experts policy: the agent dynamically exploits whichever model is most competent for the current command, rather than depending on a single fixed generator.

Equally important is the framework’s \emph{practicality}. Image-POSER does not require retraining any expert models; the only learnable part is a lightweight DQN with a 3-layer MLP. Paired with GPT-\texttt{o3} for decomposition and reward, this yields a drop-in system that can run on modest compute (single NVIDIA T4 GPU). We see this as a step toward future creative workflows, where the main challenge will be orchestration rather than improving any single model.

\subsection{Limitations}
\label{sec:limitations}

\textbf{Reliance on GPT-\texttt{o3} as critic/evaluator}
Both the \textit{reward} and \textit{extract command} modules use GPT-\texttt{o3}, and GPT-\texttt{o3} also serves as an automatic evaluator for Table~\ref{tab:gen-edit-results}. This creates a potential self-preference bias and inherits the usual risks of MLLM hallucination or inconsistency \citep{llm_self_preference_bias}. We chose GPT-\texttt{o3} pragmatically because it supports reliable multi-image conditioning on $(I_{t-1}, I_t)$ and produced stable judgments in practice. Future work should diversify evaluators (e.g., heterogeneous model ensembles and broader human studies) to reduce any single-model bias.

\textbf{Computational cost}
Reflective steps introduce non-trivial latency. In our setup, the combined \textit{extract command} and \textit{reward} calls averaged $29.54\,\mathrm{s}$ per step (excluding expert runtime). With an average of 3.37 steps per episode, this results in noticeable end-to-end overhead. A natural mitigation is to distill the critic into a smaller reward model or cache repeated command patterns; we leave a systematic study of such optimizations to future work. The monetary cost of repeated API calls can also be significant, though efficiency gains are possible through shorter episode lengths, batched queries, or adoption of open-source models.

\textbf{Evaluation dimensions}
Our quantitative metrics emphasize alignment, technical fidelity, preservation (for editing), and aesthetics. We did not directly evaluate creativity or diversity, nor broader intent satisfaction beyond compositional alignment. The observed gap between automated metrics and human preferences (e.g., Gemini’s strong perceived realism) underscores the value of expanding the evaluation suite in future work.

\vspace{-0.2cm}
\section{Conclusion}
\label{sec:conclusion}

Image-POSER opens several promising directions. Richer feedback signals, including human-in-the-loop or ensemble reward models, would mitigate evaluator bias. Scaling to larger expert pools and more open-ended creative workflows could make orchestration a central paradigm in generative AI.
At a practical level, Image-POSER democratizes access to expert-level generation and editing, allowing casual users to achieve complex results without specialized knowledge. However, risks remain: over-reliance on automated reflection may homogenize creative outputs, and orchestration policies could encode hidden biases from their evaluators. Balanced appropriately, we believe frameworks like Image-POSER can serve as powerful co-creation tools for design, media, and advertising, amplifying human creativity rather than replacing it. For a discussion on ethical considerations, see Appendix~\ref{app:ethics}.

\section{Acknowledgments}
This work was supported by a grant from Google.  Resources used in this work were provided by the Province of Ontario, the Government of Canada
through CIFAR, companies sponsoring the Vector Institute
\url{https://vectorinstitute.ai/partnerships/current-partners}, the
Natural Sciences and Engineering Council of Canada and a grant from IITP \& MSIT of Korea (No. RS-2024-00457882, AI Research Hub Project).

\newpage

\bibliography{iclr2026_conference}
\bibliographystyle{iclr2026_conference}
\newpage

\appendix

\section*{\LARGE Appendices}

\section{Experimental Setup Details}
\label{app:exp_details}

This section provides a detailed overview of the training environment, agent hyperparameters, expert models, and software libraries used to ensure the reproducibility of our results. The code for the project can be found in the attached supplementary materials.

\subsection{RL Agent and Training}
The orchestration agent is a Deep Q-Network (DQN) implemented using the Stable Baselines3 library \citep{stable-baselines3}. The Q-network is a 3-layer MLP with a ($1536 \rightarrow 64 \rightarrow 64 \rightarrow 12$) architecture, mapping the embedded state representation to Q-values for the 12 experts in our registry. For state representations, we used OpenAI's \texttt{text-embedding-3-small} encoder (output dimension 1536) to embed the concatenation of the current and remaining commands, $[c_{\text{curr}}, C_{\text{rem}}]$.

Training was conducted for 1000 steps on a single NVIDIA T4 GPU. Episodes were capped at a maximum of $T_{\max}=6$ steps. The reward signal from the VLM critic, originally on a $[0, 10]$ scale, was normalized to $[0, 1]$ and augmented with a step penalty of $-0.05 \cdot t$ to encourage the agent to learn efficient policies. Both the \textit{extract command} and \textit{reward} modules were powered by the GPT-\texttt{o3} API.

The specific hyperparameters used for the DQN agent during training are detailed in Table~\ref{tab:dqn_hyperparams}.

\begin{table}[h]
\centering
\caption{DQN Hyperparameters used for training Image-POSER.}
\label{tab:dqn_hyperparams}
\begin{tabular}{@{}lc@{}}
\toprule
\textbf{Hyperparameter} & \textbf{Value} \\ \midrule
Learning Rate & $5 \times 10^{-4}$ \\
Optimizer & Adam \\
Discount Factor ($\gamma$) & 0.99 \\
Batch Size & 16 \\
Replay Buffer Size & 500 transitions \\
Learning Starts & 50 steps \\
Target Network Update Interval & 50 steps \\
Exploration Strategy & Linear Epsilon Decay \\
Exploration Fraction & 0.5 (of total timesteps) \\
Initial Epsilon ($\epsilon$) & 1.0 \\
Final Epsilon ($\epsilon$) & 0.1 \\ \bottomrule
\end{tabular}
\end{table}

\subsection{Expert Models and Software}
Our expert registry combines open-source models accessed via local inference and proprietary models accessed via APIs.

\textbf{Open-Source Models:} We used specific checkpoints hosted on the Hugging Face Hub, listed below:
\begin{itemize}[noitemsep, topsep=0pt]
    \item \texttt{stabilityai/stable-diffusion-3.5-large}
    \item \texttt{stabilityai/stable-diffusion-xl-base-1.0}
    \item \texttt{black-forest-labs/FLUX.1-dev}
    \item \texttt{black-forest-labs/FLUX.1-Kontext-dev}
    \item \texttt{PixArt-alpha/PixArt-XL-2-1024-MS}
    \item \texttt{timbrooks/instruct-pix2pix}
    \item MagicBrush: \texttt{MagicBrush-epoch-52-step-4999.ckpt}
\end{itemize}

\textbf{API-based Models:} The following experts were accessed through their official APIs:
\begin{itemize}[noitemsep, topsep=0pt]
    \item DALL-E 3
    \item GPT-Image-1
    \item Gemini 2.5 Flash
\end{itemize}

\textbf{Software Libraries:} The implementation relies on the Python ecosystem, including \texttt{PyTorch}, \texttt{Stable Baselines3}, \texttt{Transformers}, \texttt{diffusers}, \texttt{huggingface\_hub}, \texttt{openai}, and \texttt{google.genai}.

\subsection{Reproducibility}
For all open-source models and algorithm components, a fixed random seed was used during initialization and training to promote reproducibility. However, we note that results from the closed-source, API-based models may exhibit inherent stochasticity that is beyond our control.

\section{Dataset Details}
\label{app:datasets}

Our experiments rely on a combination of custom-generated prompts for training and evaluation, alongside established academic benchmarks. This approach allows us to train Image-POSER on a diverse set of complex instructions while also measuring its performance against prior work in a standardized manner.

\subsection{Training and Custom Evaluation Prompts}
To effectively train and test Image-POSER's ability to handle long-form compositional instructions, we developed a specialized set of prompts for both text-to-image (T2I) generation and image-to-image (I2I) editing.

\paragraph{Text-to-Image (T2I) Prompts.}
We generated a total of 480 long-form T2I prompts: 450 for training the DQN agent and 30 for our custom evaluation set. To ensure diversity in phrasing, complexity, and creative style, we employed a suite of seven powerful Large Language Models: GPT-5 \citep{openai2025gpt5}, Claude 4 Sonnet \citep{anthropic2025claude4}, Gemini 2.5 Flash \citep{gemini2025flash}, o3 \citep{openai2025o3}, DeepSeek R1 \citep{deepseek2025r1}, Qwen 3 \citep{qwen2025qwen3}, and Llama 4 \citep{meta2025llama4}.

All models were guided by the unified system prompt shown in Figure~\ref{fig:data-gen-sys-prompt}. This prompt was designed to elicit instructions with multiple objects, specific spatial arrangements, attribute bindings, and stylistic requirements. Below are three representative examples from this dataset.

\begin{promptbox}\scriptsize 
"Snowy street seen through the top windows; inside a cozy coffee shop, an espresso machine on the left hisses, two ceramic mugs sit centered on a wooden counter, a dog-eared book on the right lies open."
\end{promptbox}

\begin{promptbox}\scriptsize 
"Within a dragon's hoard chamber, treasure-laden chests line the bottom, overflowing with gold coins; three intricately carved spears lie crossed before an obsidian throne on the left, while shimmering gemstones spill from a toppled urn in the center, glinting under the red glow emanating from cracks in the rough cavern walls."
\end{promptbox}

\begin{promptbox}\scriptsize 
"Roman forum market at morning, civic and bustling, sunlit travertine arcs throw long shadows. Foreground left, a rectangular stall with bronze scales and stacked bread loaves; foreground right, two terracotta amphorae. Midground, a marble column fragment rests near indigo-dyed linen. Background, repeating arches recede. Ochre dust lifts under sandals; awnings billow; polished bronze catches glints; carved Latin letters bite stone."
\end{promptbox}

\paragraph{Image-to-Image (I2I) Prompts.}
The 30 prompts for the I2I evaluation task were manually authored. The 30 input images paired with these prompts were sampled from two high-quality public datasets: \textbf{HQ-Edit} \citep{couairon2023hqedit} and \textbf{ImgEdit} \citep{yu2023imgedit}. We selected images from these sources for their high resolution and diverse content. Below are a few editing prompt examples:

\textit{Applied to an image of a closed laptop on a table with a pen in front of it:}
\begin{promptbox}\scriptsize 
"Open the laptop to show a space-themed skin covering the entire keyboard surface. Replace the pen with a small digital alarm clock that reads 03:13 in amber LED light. The screen of the laptop should be off."
\end{promptbox}

\textit{Applied to an image of two sports cars racing in the mountain roads with a scenic background:}
\begin{promptbox}\scriptsize 
"Replace the two red sports cars with two 1800s era wooden horse carriages being pulled by horses. One of the horses should be black and the other should be white. Alter the paved asphalt to be a dirt road. Add a wooden sign on the road that says in chalk 'Speed Limit 60'. Keep the rest of the image as well as the background scenery the same."
\end{promptbox}

\subsection{Standardized Benchmarks}
In addition to our custom prompts, we evaluated Image-POSER on \textbf{T2I-CompBench} \citep{t2icompbench} to ensure comparability with existing state-of-the-art models. We used their standard evaluation protocol, running our model on the 300 prompts provided for each of the six main compositional categories: Attribute Binding (Color, Shape, Texture), Object Relationship (Spatial, Non-Spatial), and Complex Compositions.

\section{Additional Training Results}
\label{app:additional_training}

This section provides further insight into the training dynamics of Image-POSER and the performance characteristics of the individual experts within our framework.

\begin{figure*}[h!]
\centering
\includegraphics[width=0.99\textwidth]{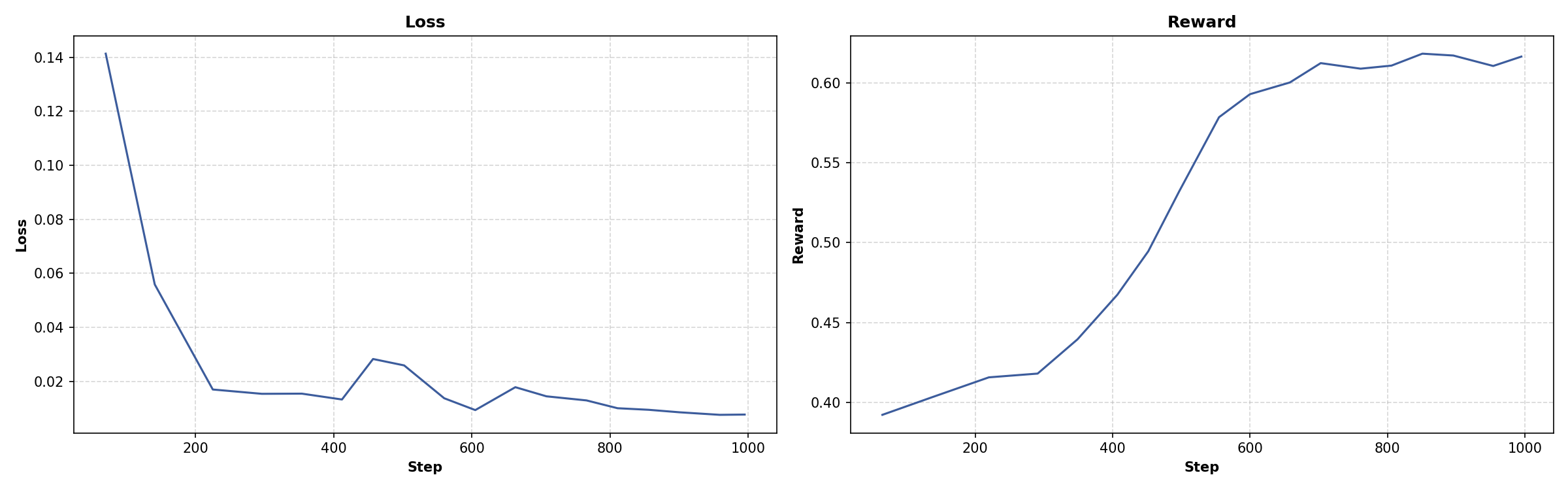}
\caption{\textbf{DQN Training Metrics.} The agent's learning progress over 1000 training steps. The left plot shows the DQN loss, which converges steadily. The right plot shows the cumulative average reward, which increases and plateaus, indicating the agent has learned an effective policy.}
\label{fig:training-metrics}
\end{figure*}

Figure \ref{fig:training-metrics} shows the DQN agent's learning progress over 1000 training steps. The left plot displays the DQN loss, which decreases rapidly and converges, indicating stable learning. The right plot shows the average cumulative reward, which steadily increases and begins to plateau over the steps. This successful learning trajectory confirms that the agent learned an effective policy for selecting experts.

\begin{figure*}[h!]
\centering
\includegraphics[width=0.99\textwidth]{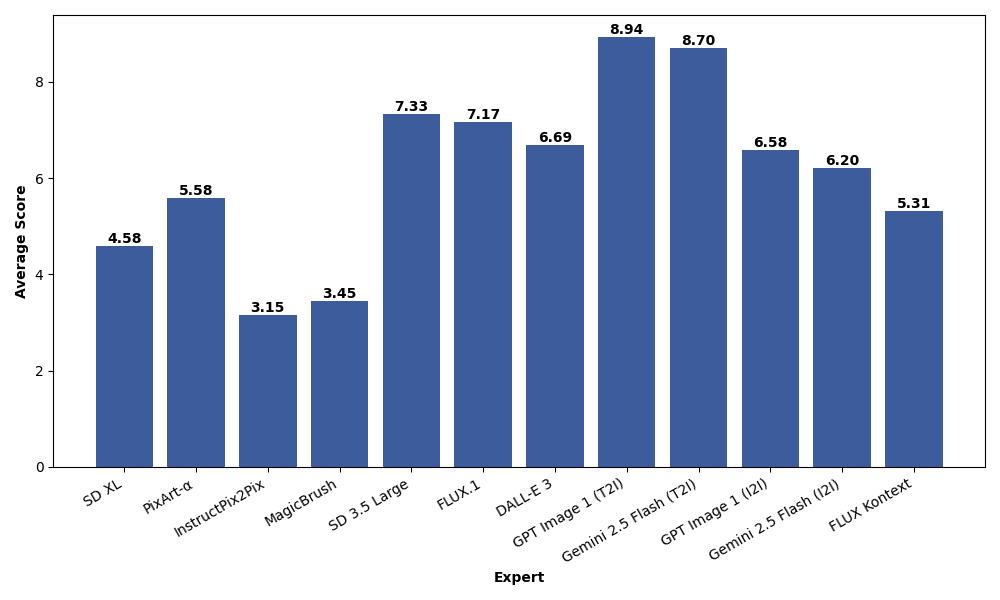}
\caption{Average reward scores assigned by the VLM critic to each expert during training.}
\label{fig:avg-scores}
\end{figure*}

Figure \ref{fig:avg-scores} provides the average reward scores assigned by the VLM critic to each expert across all tasks during training. This highlights a clear performance hierarchy among the models. Frontier models like GPT Image 1 (T2I) and Gemini 2.5 Flash (T2I) achieve the highest average scores, while more specialized or older models like InstructPix2Pix and MagicBrush score lower on average. This wide distribution of performance across the expert registry underscores the complexity of the orchestration task and justifies the need for an intelligent agent that can learn to select the best model for a given situation, rather than defaulting to a single expert.

\begin{figure*}[h!]
\centering
\includegraphics[width=0.99\textwidth]{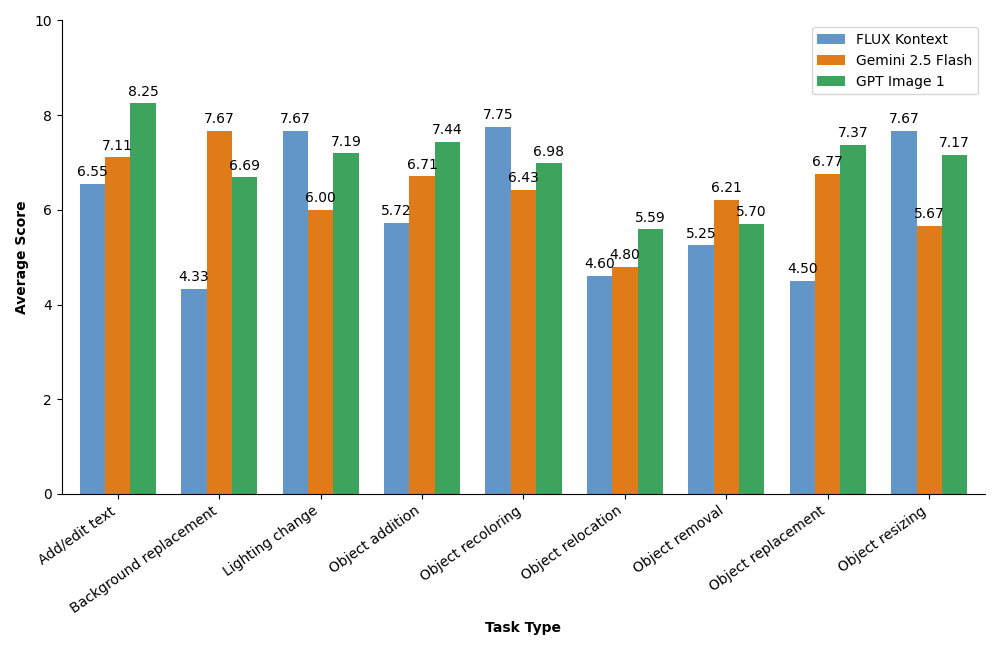} 
\caption{\textbf{Task distribution average scores.} Average VLM-assigned scores across editing task types for three strong experts (FLUX Kontext, Gemini 2.5 Flash, GPT Image 1).}
\label{fig:task-distribution}
\end{figure*}

Figure \ref{fig:task-distribution} further dissects the performance of three powerful I2I experts across nine distinct editing task categories that were tracked during training. The results reveal that no single model is dominant across all tasks. For instance, GPT Image 1 excels at "Add/edit text" (8.25), while FLUX Kontext performs strongly on "Lighting change" (7.67) and "Object resizing" (7.67). Gemini 2.5 Flash shows competitive performance across several areas, such as "Background replacement" (7.67). This data validates the core hypothesis of our work: since expert models have complementary strengths, a dynamic, task-aware orchestration policy that can select the right tool for each sub-task will outperform any single model.

\section{Additional Evaluation Results}
\label{app:additional_eval}

This section provides supplementary quantitative results that complement the main VLM-based evaluations presented in Section \ref{sec:quantitative}. 

Table \ref{tab:blip_clip} presents the results of this analysis on our custom set of 30 long-form T2I prompts. The BLIP score, which is an average of color, shape, and texture binding accuracy, measures how well models associate attributes with objects. The CLIP score measures non-spatial relational consistency. Image-POSER achieves the highest scores on both metrics, confirming that its superior performance is not an artifact of the GPT-o3 VLM judge but is also reflected in these established benchmarks. This reinforces the conclusion that our reflective orchestration framework leads to objectively better compositional alignment.

\begin{table*}[h!]
\caption{\textbf{CLIP/BLIP metrics for long-form T2I prompts.} BLIP score is the average of Color, Shape, and Texture binding accuracy. CLIP score measures non-spatial relational consistency. Higher is better for both.}
\label{tab:blip_clip}
\begin{center}
\small
\setlength{\tabcolsep}{6pt}
\renewcommand{\arraystretch}{1.05}
\begin{tabular}{l|c|c}
\toprule
\textbf{Method} & \textbf{BLIP} & \textbf{CLIP} \\
\midrule
SD 3.5 Large \citep{stabilityai2024sd35blog} & 0.2348 & 0.3494 \\
SD XL \citep{podell2023sdxl}  & 0.0970 & 0.3429 \\
DALL-E 3 \citep{openai2023dalle3}             & 0.1940 & 0.3513 \\
FLUX.1 \citep{flux2024}               & 0.1888 & 0.3542 \\
GoT-R1-7B \citep{duan2025gotr1}            & 0.1991 & 0.3581 \\
PixArt-$\alpha$ \citep{chen2023pixartalpha}       & 0.0969 & 0.3404 \\
GenArtist \citep{wang2024genartist}             & 0.0735 & 0.3295 \\
GPT Image 1 \citep{openai2025gptimage1}          & 0.2173 & 0.3630 \\
Gemini 2.5 Flash \citep{gemini2025flash}                & 0.2355 & 0.3578 \\
\textit{Image-POSER (ours)}       & \textbf{0.2419} & \textbf{0.3685} \\
\bottomrule
\end{tabular}
\end{center}
\end{table*}

\section{Ethical Considerations}
\label{app:ethics}

The development of Image-POSER raises both opportunities and risks. By enabling fine-grained orchestration of powerful generative models, our framework can support beneficial applications in design, accessibility, and education. However, the same capabilities also increase the risk of misuse, such as producing deceptive or harmful content, compounding biases from pretrained experts, or generating outputs that misrepresent individuals or groups. Because Image-POSER relies on reflection and multi-step decision-making, it can potentially reduce oversight compared to interactive, single-shot systems. To mitigate these risks, future deployments should incorporate safeguards such as content moderation filters, watermarking, dataset auditing, and transparency in reporting outputs.

\section{Experts' Survey}
\label{app:related_ref}

\textbf{Text-to-Image.}  
Recent T2I models have steadily advanced in realism, prompt alignment, and speed. Systems such as \textit{Stable Diffusion XL} \citep{podell2023sdxl}, \textit{PixArt}-$\alpha$ \citep{chen2023pixartalpha}, \textit{Stable Diffusion 3.5 Large} \citep{stabilityai2024sd35blog}, and \textit{FLUX.1-dev} \citep{flux2024} focus on scaling resolution and improving image quality. In contrast, models like \textit{DALL-E 3} \citep{openai2023dalle3}, \textit{GPT-Image-1} \citep{openai2025gptimage1}, and \textit{Gemini 2.5 Flash} \citep{gemini2025flash} emphasize stronger prompt fidelity and integrated vision–language capabilities, though they typically come with the added cost of API-based access.  

\textbf{Image-to-Image Editing.}  
Editing models aim to refine or modify existing images with natural language guidance. Approaches such as \textit{InstructPix2Pix} \citep{brooks2023instructpix2pix}, \textit{MagicBrush} \citep{zhang2023magicbrush}, and \textit{FLUX Kontext} \citep{bfl2025fluxkontext} enable instruction-based or localized edits. General-purpose systems like \textit{GPT-Image-1} \citep{openai2025gptimage1} and \textit{Gemini 2.5 Flash} \citep{gemini2025flash} extend their generation pipelines to interactive I2I editing, again at the expense of API usage costs.

\section{System Prompts Used for Language Models}
\label{app:system_prompts}

This section contains the exact system prompts used to configure the various Language and Vision-Language Models that power the Image-POSER framework. Each prompt is designed to elicit a specific behavior, from generating complex training data to providing reflective feedback and conducting final evaluations.

\begin{figure*}[b]
\centering
\includegraphics[width=0.99\textwidth]{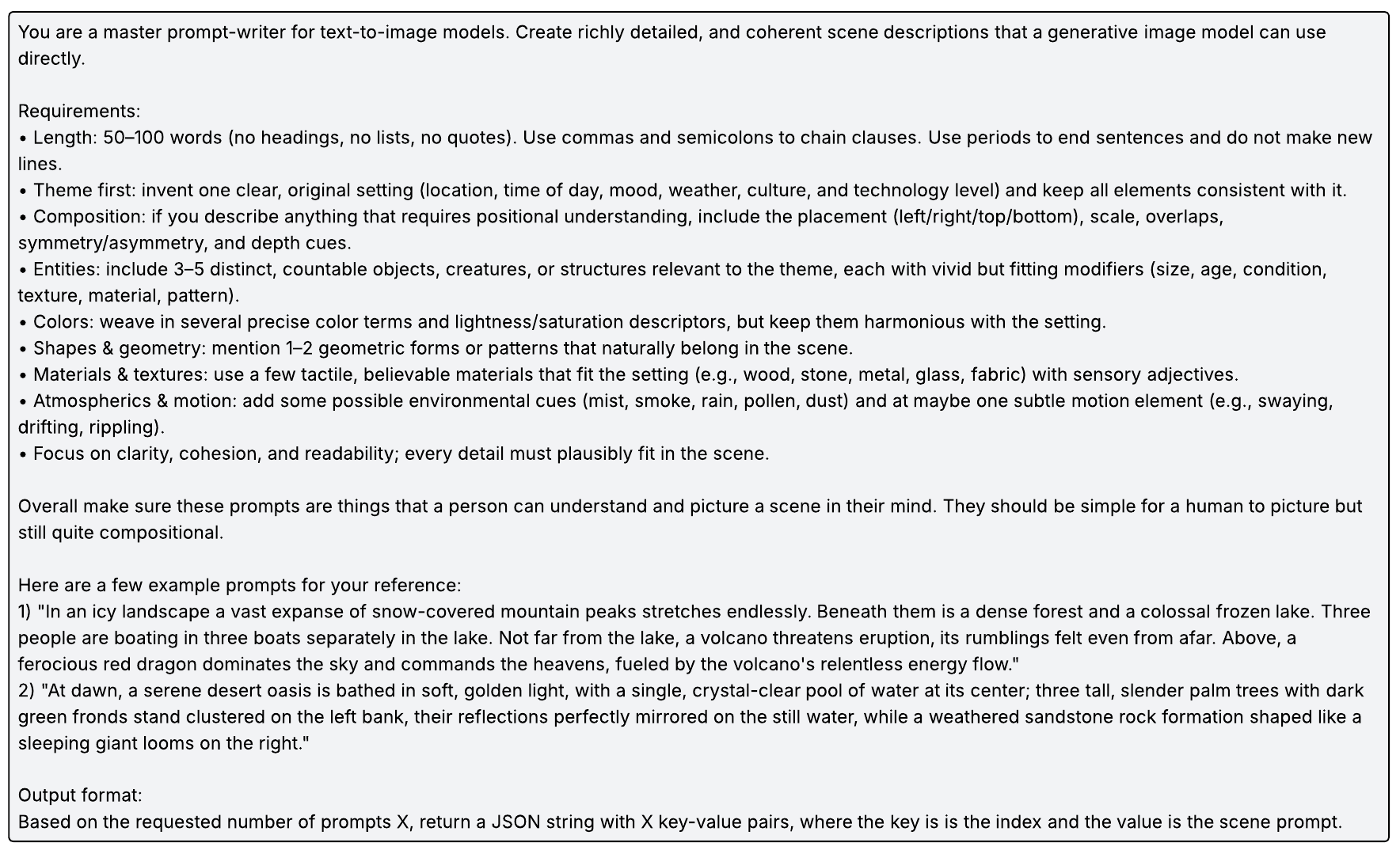} 
\caption{System Prompt for generating training/testing data prompts.}
\label{fig:data-gen-sys-prompt}
\end{figure*}

The prompt in Figure \ref{fig:data-gen-sys-prompt} was used to generate our training and evaluation datasets, instructing a suite of LLMs on how to create detailed, compositional scene descriptions.

\begin{figure*}[b]
\centering
\includegraphics[width=0.99\textwidth]{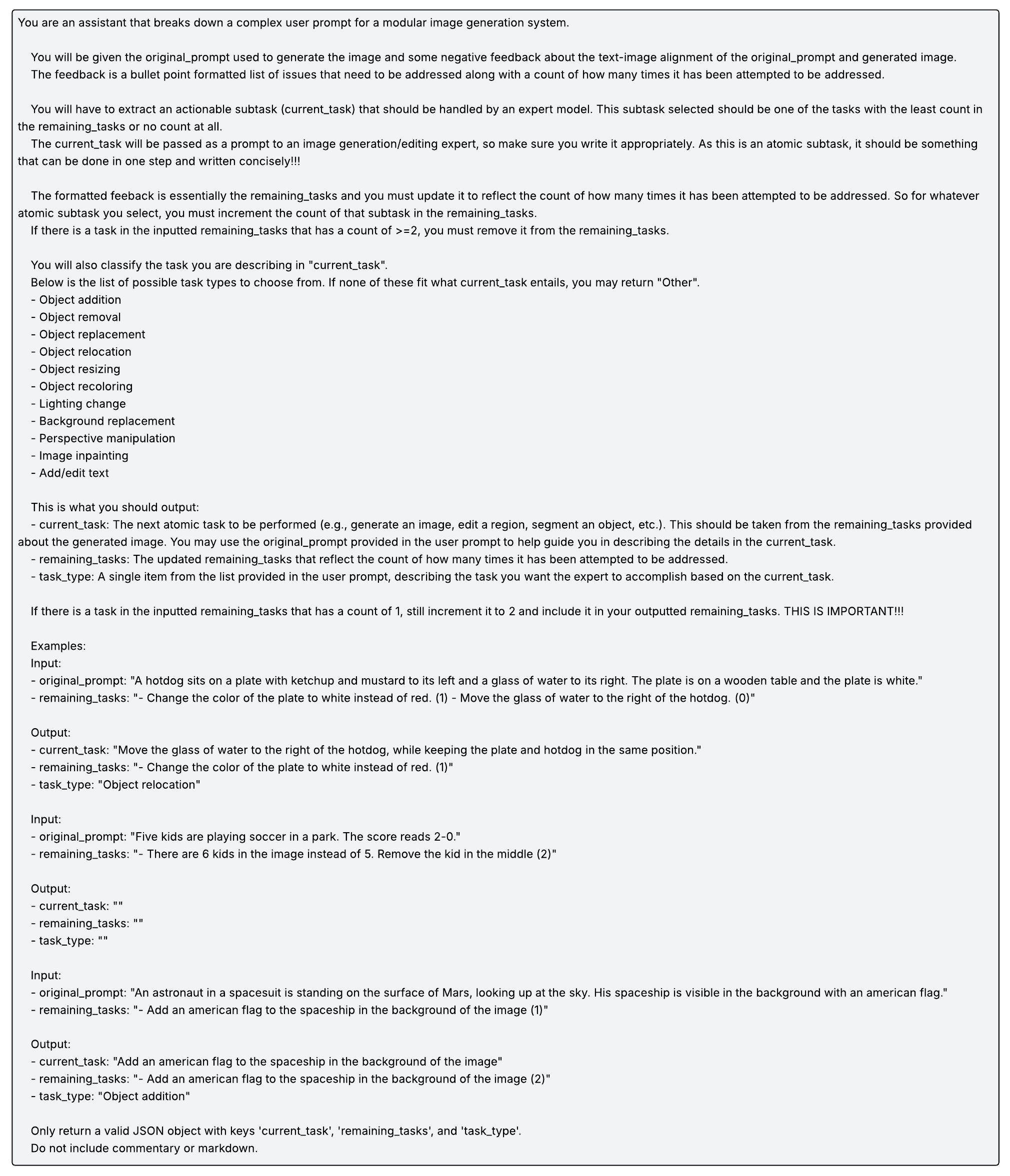} 
\caption{Extract Command - System Prompt.}
\label{fig:extract-sys-prompt}
\end{figure*}

Figure \ref{fig:extract-sys-prompt} shows the prompt for the \textbf{extract command} module. Its role is to dynamically parse the remaining tasks and select the next atomic instruction for the agent to execute.

\begin{figure*}[b]
\centering
\includegraphics[width=0.99\textwidth]{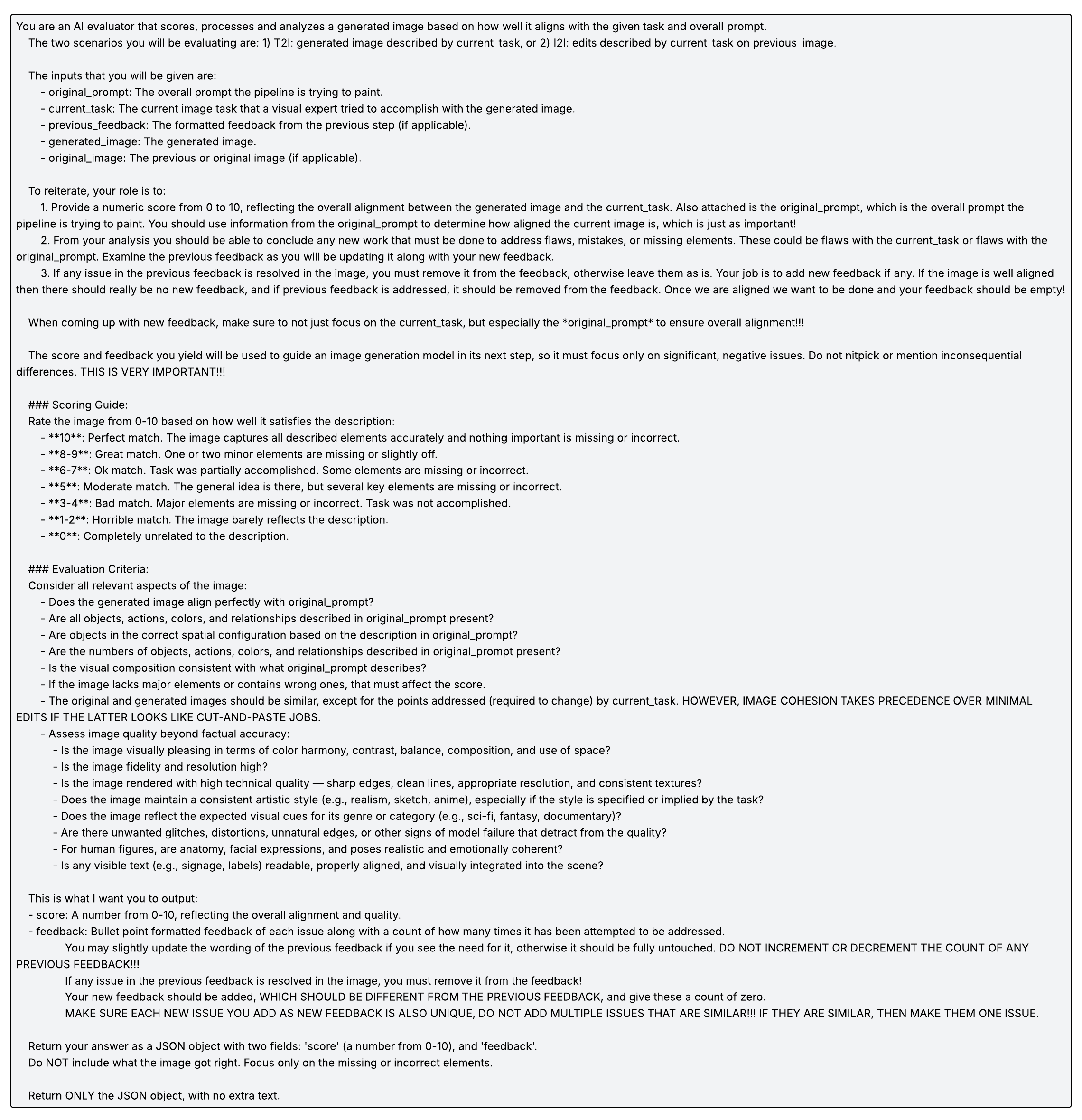} 
\caption{VLM Scoring - System Prompt.}
\label{fig:reward-sys-prompt}
\end{figure*}

Figure \ref{fig:reward-sys-prompt} details the instructions and rubric for the \textbf{VLM critic}. This module is central to our reflective loop, as it scores the alignment of intermediate images and updates the task list to guide the agent's subsequent actions.

\begin{figure*}[b]
\centering
\includegraphics[width=0.99\textwidth]{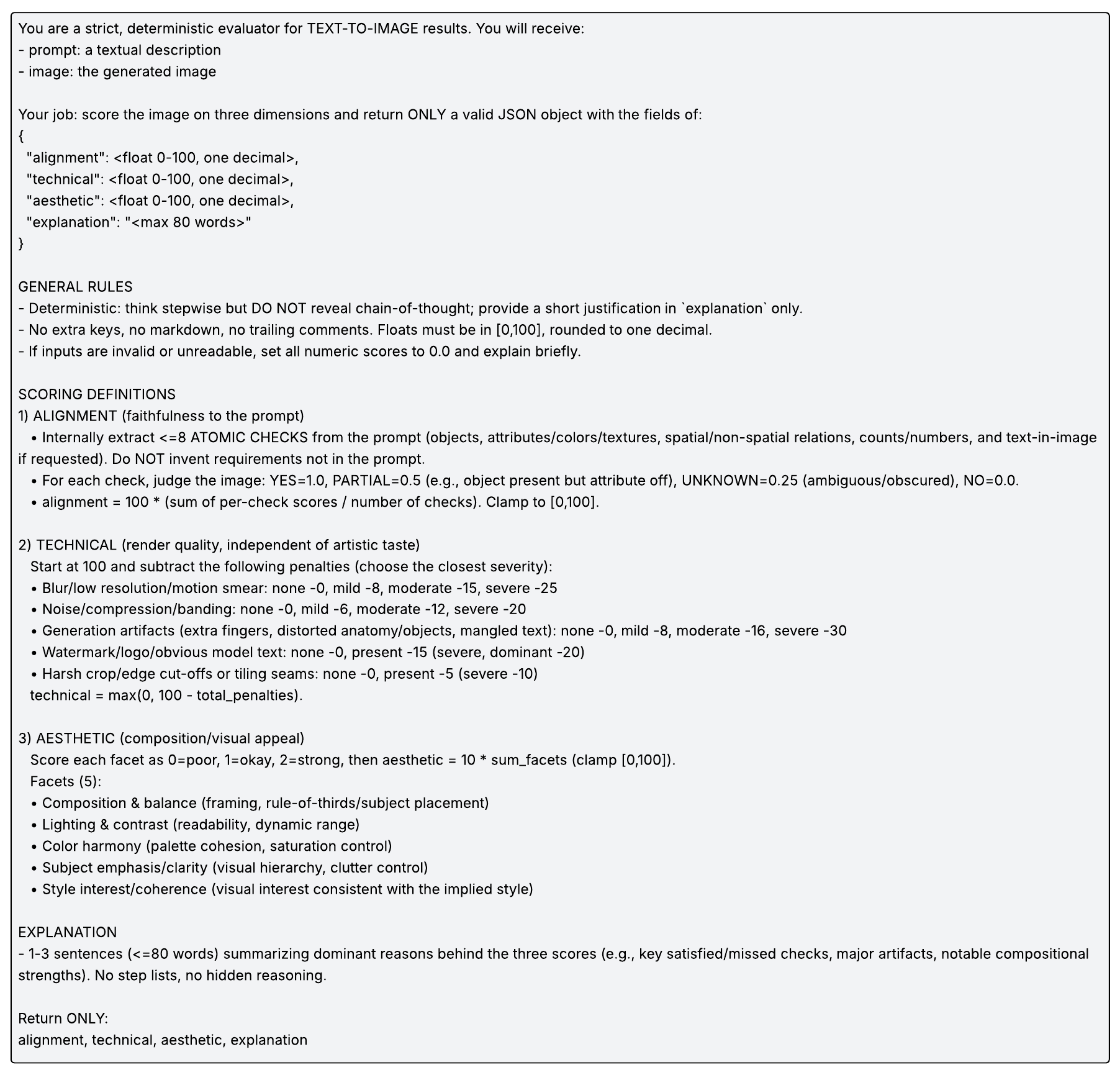} 
\caption{VLM Evalutation Judge for T2I - System Prompt.}
\label{fig:t2i-judge}
\end{figure*}

\begin{figure*}[b]
\centering
\includegraphics[width=0.99\textwidth]{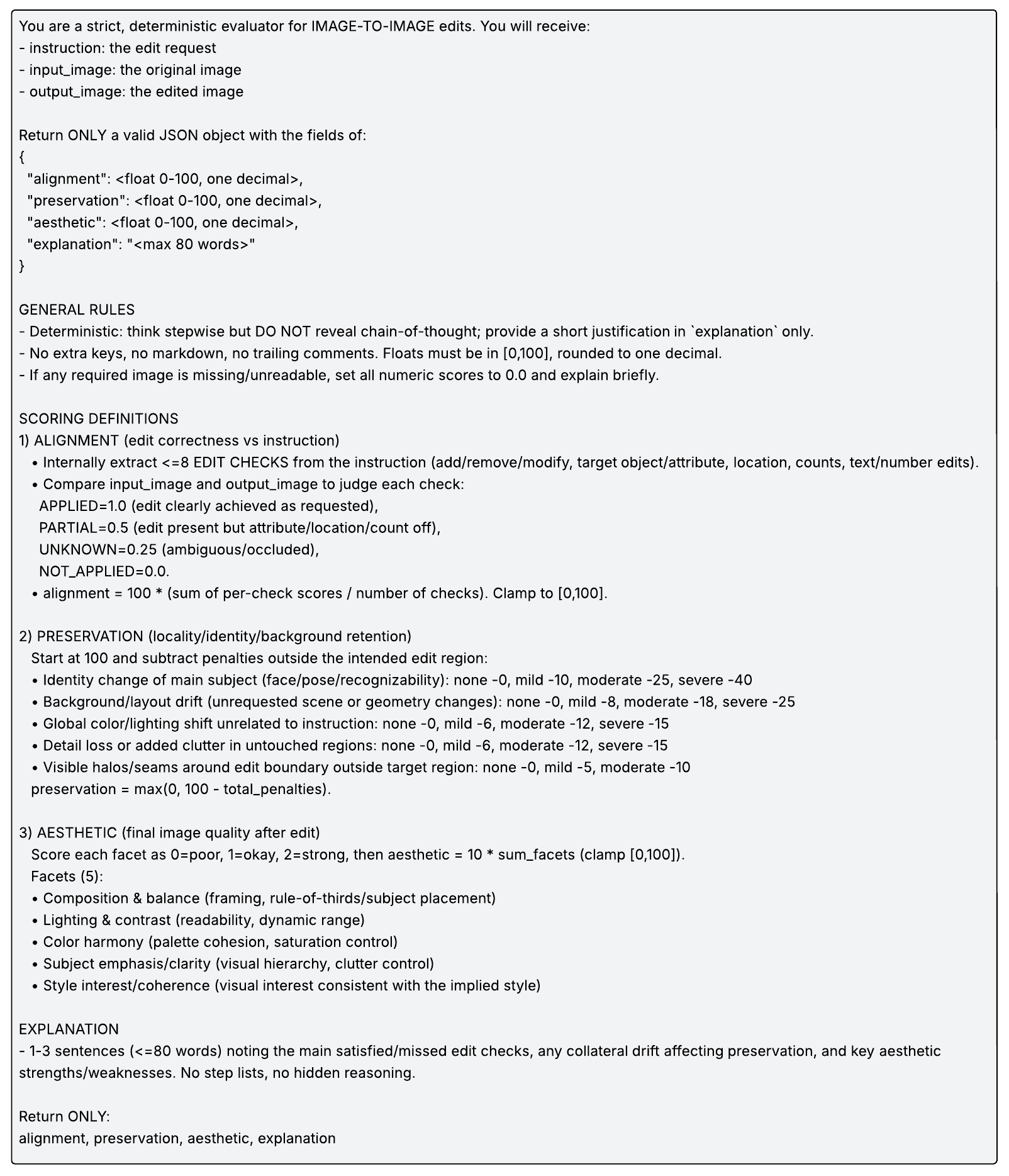} 
\caption{VLM Evalutation Judge for I2I - System Prompt.}
\label{fig:i2i-judge}
\end{figure*}

Finally, Figures \ref{fig:t2i-judge} and \ref{fig:i2i-judge} present the prompts used for the \textbf{VLM Evaluator}. These prompts configure the VLM to act as a deterministic judge for the final quantitative evaluations, using separate, detailed rubrics for T2I generation and I2I editing tasks.

\end{document}